\documentclass[a4paper, 10pt, conference]{ieeeconf}      % Use this line for a4
\IEEEoverridecommandlockouts                              % This command is only
\usepackage{graphics} % for pdf, bitmapped graphics files
\usepackage{epsfig} % for postscript graphics files
\usepackage{mathptmx} % assumes new font selection scheme installed
\usepackage{times} % assumes new font selection scheme installed
\usepackage{amsmath} % assumes amsmath package installed
\usepackage{amssymb}  % assumes amsmath package installed
\usepackage{multirow}
\usepackage{bm}
\usepackage{xcolor}
\usepackage{here}
\RequirePackage{algorithm2e}
\usepackage[subrefformat=parens]{subcaption}
\title{\LARGE \bf
Error Identification and Recovery in Robotic Snap Assembly
}

\author{Yusuke Hayami$^{1}$, Weiwei Wan$^{1}$, Keisuke Koyama$^{1}$, Peihao Shi$^1$, Juan Rojas$^2$ and Kensuke Harada$^{1}$% <-this % stops a space

\thanks{$^{1}$Graduate School
of Engineering Science, Osaka University; 1-3 Machikaneyama-
cho, Toyonaka, Osaka 560-8531, JAPAN
        {\tt\small harada@sys.es.osaka-u.ac.jp}}%
\thanks{$^2$ Chinese University of Hong Kong}
}

\begin{document}

\maketitle
\thispagestyle{empty}
\pagestyle{empty}

%%%%%%%%%%%%%%%%%%%%%%%%%%%%%%%%%%%%%%%%%%%%%%%%%%%%%%%%%%%%%%%%%%%%%%%%%%%%%%%%
% ========
%  Body
% ========

 \begin{abstract}
Existing methods for predicting robotic snap joint assembly cannot predict failures before their occurrence. To address this limitation, this paper proposes a method for predicting error states before the occurence of error, thereby enabling timely recovery.  Robotic snap joint assembly requires precise positioning; therefore, 
even a slight offset between parts can lead to assembly failure.  
To correctly predict error states, we apply functional principal component analysis (fPCA) to 6D force/torque profiles that are terminated before the occurence of an error. 
The error state is identified by applying a feature vector to a decision tree, wherein the support vector machine (SVM) is employed at each node. If the estimation accuracy is low, we perform additional probing to more correctly identify the error state. Finally, after identifying the error state, a robot performs the error recovery motion based on the identified error state. 
Through the experimental results of assembling plastic parts with four snap joints, we show that the error states can be correctly estimated and a robot can recover from the identified error state. 

%
% 本稿では，
% ロボットによるスナップアセンブリの作業途中に自動でエラー状態から復帰することを目的とし，作業の成功，及び複数の失敗パターンを学習することでエラーの発生を予測し，エラーパターンの識別を行う手法，さらに，識別結果に基づいた適切なエラーリカバリ行動を生成することで作業へ復帰するエラーリカバリシステムを提案する．
% スナップアセンブリにおける従来のエラーパターン識別手法では，識別のために作業の終了時までの多量な力データを必要とするため，作業の途中で作業結果を識別することは困難であり，エラー状態からの復帰が不可能である．
% そこで本稿では，特徴量抽出手法として関数主成分分析を用いることで作業途中までのデータのみからパターン識別に有用なデータを獲得し，サポートベクターマシン(SVM)による機械学習を通じてエラーパターンを識別する手法を提案する．その後，識別結果に基づいて，作業部品が破損しないようなリカバリ行動によって作業への復帰を行うことで，エラーリカバリシステムを構築する．
% 本稿では，まず取得データから特徴量を抽出する手法について述べた後で，エラーパターンの分類器の構築手法について述べる．
% その後，作業途中でのエラーパターン識別実験を行い，エラーリカバリ行動の例を示すことで，
% 作業途中でのエラーパターン識別手法，及び，それを用いたエラーリカバリシステムの実現可能性について検討する．
% それぞれの手法に対する評価実験の結果，提案手法の有用性が示唆される．
%
\end{abstract}
 \section{Introduction}
\label{1}
\begin{figure}[t]
  \begin{center}
          \includegraphics[width=\linewidth]{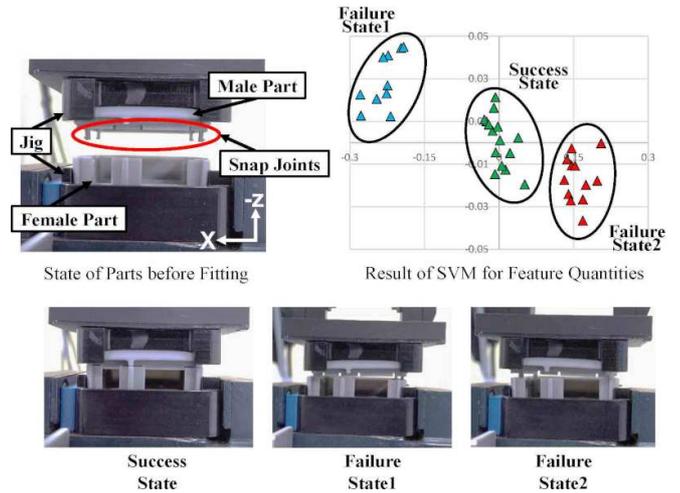}
          \caption{Identification of success/error states in robotic snap assembly}
          \label{SVM overview}
        \end{center}
\end{figure}

Robotic product assembly has recently been introduced to some production processes. 
However, we often encounter products that are difficult for robots to assemble with a high success rate.  
In this context, this research focuses on the robotic assembly of plastic parts that include snap joints, hereafter called robotic snap assembly, which is often difficult owing to the elasticity of parts. 
Plastic parts with snap joints are easy to assemble but highly difficult to disassemble; consequently, it becomes crucial to correctly predict assembly failure before the assembly actually fails. Moreover, the deviation in the position of the part, which causes the assembly failure, is usually minute; thus, it becomes difficult for a vision sensor to predict assembly failure. To address this problem, this research proposes a method for error identification and recovery in robotic snap assembly based on 6D force/torque sensor information attached at the wrist. An overview of our proposed method is shown in Fig. \ref{SVM overview}. 

% スナップアセンブリとは，スナップジョイントと呼ばれる突起を有する部品を嵌め合う組立作業であるが，その際，作業部品を固定する治具が経年劣化等を起こすことで，部品間にわずかな位置のずれ(以降，このずれをオフセットとよぶ)が発生し，嵌め合いが失敗することがある(図\ref{snap assembly overview})．
% このため，ロボットハンドや作業部品が破損することや，過大な力を検出してロボットが緊急停止した場合は，原因究明のために生産ラインを停止しなければならないといったことが問題点として挙げられる．また，このようなオフセット値はわずかなものであるため，ビジョンセンサ等の画像処理技術のみによって解決することは困難である．

%\begin{figure}[h]
% \begin{minipage}[b]{0.49\linewidth}
%  \centering
%  \includegraphics[width=0.83\linewidth]{picture/chap1/before_fitting_parts.eps}
%  \subcaption{State of parts before fitting}
% \end{minipage}
% \begin{minipage}[b]{0.49\linewidth}
%  \centering
%  \includegraphics[width=0.84\linewidth]{picture/chap1/success_pattern1.eps}
%  \subcaption{Successful fitting pattern}
% \end{minipage}\\
% \begin{minipage}[b]{0.49\linewidth}
%  \centering
%  \includegraphics[width=0.8\linewidth]{picture/chap1//failure_pattern2.eps}
%  \subcaption{Failure fitting pattern 1}
% \end{minipage}
% \begin{minipage}[b]{0.49\linewidth}
%  \centering
%  \includegraphics[width=0.9\linewidth]{picture/chap1/failure_pattern1.eps}
%  \subcaption{Failure fitting pattern 2}
% \end{minipage}
% \caption{Successful/failure patterns in robotic snap assembly}\label{snap assembly overview}
%\end{figure}

Thus far, a few researchers have proposed methods for identifying error states in robotic snap assembly using force/torque information
\cite{1,2,3,4,5,6,7,rojas1,rojas2,rojas3,rojas4,rojas5,rojas6,lello}. However, in all of those methods, the error was identified after completing the assembly task; consequently it becomes difficult to recover from the error state. In contrast, this paper proposes a method for identifying an error state among multiple possible error states during robotic snap assembly. To identify an error state, we apply functional principal component analysis (fPCA) to the force/torque profile of an assembly task. Furthermore, to identify an error state among multiple possible error states, we construct a decision tree wherein the classification of the force/torque profile is performed at each node with  the support vector machine (SVM) by employing the kernel function.   
In addition, if the estimation accuracy is not sufficiently high, we perform additional probing, whereby the part is moved to more correctly identify the error state. After the error state is identified, a robot retries the assembly task by slightly modifying the initial position/orientation of the part in the direction opposite to that in the identified error state. 

% スナップアセンブリにおける従来のエラーパターン識別手法では，上記のような問題に対して力覚情報を用いることで，作業が成功したのか，または失敗した場合はどの方向にずれが発生して，つまり，どのようなエラー状態で失敗したのか(以下，成否パターンと呼ぶ)といった成否パターンの識別は達成されている．しかしながら，これらの手法では，識別のために作業の終了時までの多量な力データを必要とするため，作業の途中で成否パターンを識別することは困難である．
% 実際にロボットがスナップアセンブリにおいてエラーリカバリ動作を行う場合，作業が終了するよりも前の時点でその作業の成否パターンを識別することが必要となる．そのため，作業途中での成否パターンの識別手法は，ロボットによるスナップアセンブリの自動化には必要不可欠であると考えられる．

% そこで本稿では，スナップアセンブリにおいて，作業途中で成否パターンを識別し，エラー状態であると識別された場合は，識別した成否パターンに基づいたエラーリカバリ動作を行うことで，作業への復帰を実現する
% 手法について検討する．
% 対象とするエラーリカバリシステムの構築にあたり，本研究では特に，成否パターンを分類するような分類器を構築する問題と，成否パターンの識別を作業途中で行う問題，ならびにエラーからのリカバリを行う問題に取り組む．
% 成否パターンの分類問題については，ロボットハンドに搭載された力・トルクセンサから取得される波形データから抽出される特徴量を作業結果の成否のみだけでなく，成否パターンまで細分化して分類することで分類器を構築する．本研究では，分類のためにサポートベクターマシン(以下，SVMと呼ぶ)を用いて，成否パターンに対する決定木を構築することで学習を行う．
% 成否パターンの分類の様子を
% 図\ref{SVM overview}に示す．
%\begin{figure}[t]
%  \begin{center}
%          \includegraphics[width=\linewidth]{picture/chap1/SVM_implementation.eps}
%          \caption{Classification of successful/failure patterns by SVM}
%          \label{SVM overview}
%        \end{center}
%\end{figure}

The remainder of this paper is organized as follows: A few relevant previous works are described in Section 2, the proposed method is elucidated in Section 3, and we present our experimental results in Section 4. Finally, summary and scope for further research are presented in Section 5. 

% また，作業途中での成否パターンの識別問題については，未知のスナップアセンブリに対して抽出される特徴量に対して，決定木により分類されたクラスのいずれに相当するのかを求めることで行う．ここで，本研究では未知のオフセットパターンを持つスナップアセンブリに対して作業の途中で識別を行う必要があるため，作業途中までの力・トルクデータのみを用いて成否パターンを識別しなければならない．
% この場合，識別に必要な力・トルクデータが不足することにより，識別精度が低下することが考えられる．
% そこで，取得される波形データから波形の概形情報を抽出することができれば，識別のために利用できるデータ数が制限された場合においても，識別に必要な情報が得られると予想し，時系列データからの特徴量抽出に有用な関数主成分分析を用いる．

% さらに，上記の手法に加えて本研究では，アセンブリ工程のみでは十分な識別が困難である場合に限り，
% 追加で探り動作工程を設けることで詳細な力情報を取得し，より正確な識別を行うこととする．
% 最後に，エラーリカバリ動作の生成については，
% 識別結果に対して，識別されたずれを解消する方向にロボットハンドを移動させることで行う．
% 以降，本論文の構成について述べる．
% \ref{2}章で，部品組み立て作業における，エラーの検知・識別に関する関連研究を，
% \ref{3}章では，本提案手法について述べる．
% \ref{4}章で，実機実験結果を示し，
% \ref{5}章では，実験結果の考察を行い，提案手法の有用性について検討を行う．
% 最後に，\ref{6}章で，本論文を結論付ける．

 \section{related work}
\label{2}

Robotic assembly has been researched for decades\cite{whitney1977,hogan1985,deMello1988,wilson1994}. The research on robotic assembly has been mainly done on force controlled assembly \cite{whitney1977,hogan1985}, assembly motion planning \cite{deMello1988,wilson1994,thomas2015,wan2016,dogar2015}, learning based methods \cite{thomas2019,beltran2020}, and gripper design \cite{pham}. 

Recently, the discrimination of failure states in robotic assembly tasks has been studied by some researchers such as \cite{1}\cite{2}\cite{3}\cite{4}\cite{5}\cite{6}\cite{7}. 
Rodrigues et al.\cite{1} identified assembly failure using the SVM and PCA. Rojas et al. \cite{rojas1}\cite{rojas2}\cite{rojas3}\cite{rojas4}\cite{rojas5}\cite{rojas6} identified an error state among multiple possible states using relative-change-based hierarchical taxonomy (RCBHT).  
Lello et al. \cite{lello} discriminated success/failure of snap assembly based on Bayesian structural time series. However, in all the previous studies, the force/torque profile that is obtained upon the completion of the assembly is used, and it is impossible to predict an error state among multiple possible states before the error actually occurs. 

With respect to the robotic manipulation researches on recovery from error states, some researchers \cite{Dang2013,Li2014} used tactile information to identify and recover from the error states. On the other hand, this research uses a 6D force/torque sensor attached at the wrist during a snap joint assembly to recover from the identified error state.

% ロボットによる組立作業において，作業の成否の識別を目的とした研究が行われている\cite{1-1}\cite{1-2}\cite{1-3}\cite{1-4}\cite{1-5}\cite{1-6}\cite{1-7}．Rodriguezらは，SVMや主成分分析(PCA)を用いて作業の成功データと失敗データを学習させることで，スナップアセンブリの成否を識別する手法を提案した\cite{1-1}．一方，作業の成否のみだけではなく，失敗のパターンも識別することを目的とした研究として，Rojasら\cite{rojas}\cite{rojas1}\cite{rojas2}\cite{rojas3}\cite{rojas4}\cite{rojas5}は，カンチレバー型のスナップジョイントのアセンブリを題材として，Relative-Change-Based Hierarchial Taxonomy(RCBHT)\cite{rojas}を利用して力波形の文脈を読み取ることで成否パターンを識別する手法を提案している\cite{rojas1}．また，Lelloらは，ベイジアン時系列モデルを用いることで，作業の失敗パターンモデルを推定することにより，成否パターンを識別した\cite{lello}．
% しかしながら，これらの研究では，成否パターンの識別のために作業終了時までの多量のデータを必要としており，ロボットが作業している途中で成否パターンを識別するようなことは想定されていない．このような場合，エラーリカバリ行動が間に合わず，ロボットハンドや作業部品が破損する恐れがある．
% それに対して本研究では，ロボットによるスナップアセンブリの作業途中に自動でエラー状態から復帰することを目的とし，これらエラーパターンの識別を作業途中で行う手法，さらに，識別結果に基づいた適切なエラーリカバリ行動を生成することで作業へ復帰するエラーリカバリシステムを提案する。
% %また，識別のためにアセンブリ工程に探り動作を設けている場合，全体の作業時間が遅れる課題も発生し，探り動作を必要としない成否パターン識別手法に対する需要も高まっている．

 \section{Proposed method}
\label{3}

The proposed method is composed of offline and online phases. 
In the offline phase, we collect the training data, that is, the force/torque profiles obtained by using the 6D force/toque sensor attached at the wrist during robotic snap assembly, and construct the decision tree for classifying error states. 
In the online phase, based on a given force/torque profile corresponding to a snap assembly that was terminated before the error actually occured, we predict an error state using the constructed decision tree. 
If the accuracy of prediction is not sufficiently high, we perform additional probing by moving the part to more correctly identify the error state.  
After the error state is identified, a robot tries to recover from it. The robot retries the assembly task by modifying the initial position/orientation of the part in the direction opposite to that in the identified error state. 

% 本章では，成否パターンの分類を行うための決定木を構築する問題，及び，構築された決定木を用いて成否パターンの識別を作業途中で行い，エラーリカバリ動作により作業へ復帰するためのエラーリカバリシステムの構築問題に対して説明を行う．

% 決定木の構築問題については，
% 提案手法では，様々なオフセットパターン下で取得された力・トルク波形に対して，
% 関数主成分分析によって特徴量を抽出する．
% その後，SVMにより特徴量の持つ成否パターンの分類を行う．
% ここで，後述にある通り，発生するオフセットパターンによって分類に有効な特徴量内の力・トルク成分が異なる点を考慮して，提案手法では，決定木を構築することで成否パターンを分類する．

% また，エラーリカバリシステムの構築問題については，
% 未知のオフセットパターンにて抽出される特徴量が，決定木のいずれのクラスであるかを求めることで作業途中での識別を行う．その後，識別結果に応じて発生オフセットを解消する方向にロボットハンドを移動させることでエラーリカバリ行動を生成し，エラーリカバリシステムを実現する．

\subsection{Problem Definition}
\label{3.1}
%本研究における問題設定に関して説明する．ロボットによるスナップアセンブリの概観，及び座標系を
%図\ref{setting_chap3}に示す．
%\begin{figure}[t]
%  \begin{center}
%          \includegraphics[width=0.5\linewidth]{picture/chap3/overview_snap.eps}
%          \caption{Overview of snap assembly}
%          \label{setting_chap3}
%        \end{center}
%\end{figure}

We consider the assembly of plastic parts with four snap joints, as shown in Fig. \ref{SVM overview}. 
In robotic snap assembly, a robot holds the male part, including the snap joints, and moves 
in the $+z$ (horizontally downward) direction to fit this part to the corresponding female part. During assembly, we obtain the 
6D force/torque information using a force/torque sensor attached at the robot wrist. 
We perform assembly experiment by shifting the initial position/orientation of the male part by various values. We call such deviation of initial position/orientation in multiple directions as the offset pattern. For simplicity, this research considers the offset patterns assuming the $x$-directional translation ($\Delta x$) and the rotation about the $x$-axis ($\Delta \theta_z$). 
Based on the 6D force/torque profile of a snap assembly, we determine (1) assembly success or the directional offset, that causes assembly failure, which is categorized as

\noindent
\begin{tabular}{ll}
 (2)\ $\Delta x\geq0$, &  (3)\ $\Delta x\leq0$\\
 (4)\ $\Delta \theta_z\geq0$, & (5)\ $\Delta \theta_z\leq0$\\
 (6)\ $\Delta x\geq0$,$\Delta \theta_z\geq0$, & (7)\ $\Delta x\geq0$, $\Delta \theta_z\leq0$\\
 (8)\ $\Delta x\leq0$, $\Delta \theta_z\geq0$, and & (9)\ $\Delta x\leq0$, $\Delta \theta_z\leq0$.
 \end{tabular} 
 
\noindent
We especially consider directional offsets (2) $\cdots$ (9) as the error states. Thus, this research predicts the assembly success and error states (1) $\cdots$ (9) using the assembly force/torque profile. 

% まず，スナップアセンブリ工程については，
% 突起部分を有する部品がロボットハンドで把持された状態から，
% 作業部品間にオフセットを発生させることで成否パターンを再現する．
% その後，
% 嵌め合い対象部品に向かってロボットハンドを
% 下($+z$)方向
% に降ろすことで達成されることとする．
% 以下，スナップアセンブリの嵌め合いが完了するより前の時点までロボットハンドを
% 降ろす工程をスナップ工程と呼ぶ．
% ここで，実際に成否パターンの識別を行う際，成否パターンが複雑である場合においては，スナップ工程から取得される力情報のみでは十分な識別精度が得られないことが想定される．
% そこで，スナップ工程のみでは正しい識別が困難であると判定された場合に限り，補助的にロボットハンドに探り動作を設けることとする．
%%%%%%%%%%%%%%%%%%%%%%%%%%%%%%%%%%%%%%%%%%%%%
%% これにより，追加で探り動作時の力情報を取得し，成否パターンの識別に用いる．
%% 提案手法では，探り動作はロボットハンドを数ミリ程度真上($-z$)方向に移動させた後，$\pm{x}$方向に数ミリ程度ロボットハンドを往復させる動作とする．以下，$+x$軸方向，及び$-x$軸方向への探り動作をそれ%ぞれ右探り動作工程，左探り動作工程と呼ぶ．
%%%%%%%%%%%%%%%%%%%%%%%%%%%%%%%%%%%%%%%%%%%%%%%%%%

% また，作業部品に発生するオフセットの種類については，$x$軸方向の並進オフセットと$z$軸まわりの回転オフセットの2種類を想定する(図\ref{setting_chap3}参照)．以下，$x$軸方向の並進オフセット値，及び$z$軸まわりの回転オフセット値をそれぞれ
% $\Delta x[{\mathrm{mm}}]$，$\Delta \theta_z[^\circ]$とする．

% 上記の想定から，本研究においては，以下の成功パターン(1)，及び(2)$\sim$(9)のオフセットを持つ失敗パターンを成否パターンとして定義する．\\
% (1)成功パターン\\
% (2)$\Delta x\geq0[{\mathrm{mm}}]$，
% (3)$\Delta x\leq0[{\mathrm{mm}}]$\\
% (4)$\Delta \theta_z\geq0[^\circ]$，
% (5)$\Delta \theta_z\leq0[^\circ]$ \\
% (6)$\Delta x\geq0[{\mathrm{mm}}]$，$\Delta \theta_z\geq0[^\circ]$，
% (7)$\Delta x\geq0[{\mathrm{mm}}]$，$\Delta \theta_z\leq0[^\circ]$\\
% (8)$\Delta x\leq0[{\mathrm{mm}}]$，$\Delta \theta_z\geq0[^\circ]$，
% (9)$\Delta x\leq0[{\mathrm{mm}}]$，$\Delta \theta_z\leq0[^\circ]$

\subsection{Construction of decision tree} 

To classify error states based on the force/torque information, we define the feature vector composed of the principal component score obtained via functional principal component analysis (fPCA). Using the obtained feature vector, we construct the decision tree to classify error states. 

% スナップアセンブリにおける成否パターンを作業途中で識別するにあたり，作業途中までのデータのみから識別に有用な情報を取得する必要がある．そこで，提案手法では，関数主成分分析を用いて関数主成分得点を特徴量として抽出する．関数主成分分析により，力波形の概形情報が取得されるため，作業途中までのデータのみであっても情報の損失を低減することが可能となる．
% 次に，抽出された特徴量を学習することで，成否パターンを分類する決定木を構築する．
% ここで，スナップアセンブリにおいては上述にある通り，分類すべき成否パターンは明示的に与えることが可能であることや，成否パターン数についても高々10種類程度であることが想定される．
% そこで提案手法では，SVMを用いて，力情報から抽出された特徴量に対して，教師データとして定義された成否パターンに基づく分類を行うことで上記を実現する．
% 本節では，これら手順に関して順を追って説明していく．

\subsubsection{Data collection}
\label{3.2_sec:obtain_data}

We collect the force/torque profile during the assembly task to construct the decision tree for the purpose of identifying error states. 
We assume that there are $N$ offset patterns with regard to the initial position/orientation of the male part. 
Then, the male part is moved in the $+z$ direction so that it can be fit to the female part to check whether the assembly succeeded or not. 

% 本節では，スナップアセンブリ時の力・トルクデータを取得する．
% 作業部品間に様々なオフセットを与えながら，
%%%% スナップ工程，右探り動作工程，左探り動作工程のそれぞれに分けて
%%%%% ロボットハンドに搭載されている6次元の力・トルクセンサから計測された力・トルク成分波形を取得する．

\subsubsection{Feature extraction}
\label{3.3_sec:feature_extraction}

In this subsection, we explain the manner in which the feature quantities are defined using the 6D force/torque profile. 
Let the time trajectory of each force/torque component corresponding to the $i$-th offset pattern be $f_i(t)$. 
We apply fPCA \cite{fpca1} to $f_i(t)$.  
Among the principal component $\xi(s)$ satisfying the following equation, we collect $p$ from the one with the largest eigenvalue of $\rho$.  
%
% 本節では，\ref{3.2_sec:obtain_data}節で得られた波形データからの特徴量抽出手法について説明する．また，以下ではスナップ工程，右探り動作工程，左探り動作工程それぞれで取得された波形データに対して同様の処理を行うため，スナップ工程のみについて説明する．
% スナップ工程により，時間の関数として6次元の力・トルクの情報が得られる.ここから，波形データの特徴を良く表す複数次元の特徴量ベクトルを求め，それをSVMに適用する．ここで，本研究では作業途中での成否パターンの識別を目的としているため，抽出される特徴量は，作業途中までの力・トルクデータのみを用いた場合でも，なるべく作業終了時までの力・トルク情報が損なわれないことが必要である．そこで，特徴量抽出手法としては，関数主成分分析\cite{fpca1}\cite{fpca2}\cite{fpca3}を用いて関数主成分得点を抽出することで波形の概形情報を取得する．関数主成分分析とは，離散データに対して行う主成分分析を時系列データに応用したものである．各オフセットパターン$i=1,2,...N$において，B-spline関数を用いて関数化された各力・トルク成分$f_i(t)$を用いて主成分関数は，以下の式\ref{fpca formula}を満たす$\xi(s)$のうち，固有値$\rho$の値が大きなものから順番に$\xi$を選ぶ固有値問題に帰着される．
\begin{center}
\begin{eqnarray*}
&&\int v(s,t)\xi(s) dt=\rho\xi(t), \label{fpca formula} \\
&&\hspace*{-0.5cm}v(s,t)=\frac{1}{N}\sum_{i=1}^N (f_i(s)-\bar{f}(s))(f_i(t)-\bar{f}(t)),
\end{eqnarray*}
\end{center}
where $v(s,t)$ denotes the covariance matrix. We can now define a $p$-dimensional feature vector for each force/torque component corresponding to the $i$-th offset pattern. 
Here, if we consider a single $6p$-dimensional feature vector, the error classification may depend on some specific force/torque components. Hence, we consider six $p$-dimensional feature vectors where the SVM is separately applied to each force/torque component. 

% ここで，$v(s,t)$は共分散関数である．これにより，固有値の大きいものから順に$p$個の固有値とそれらに対応する$p$個の主成分関数が得られる． 

% 提案手法では，\ref{3.2_sec:obtain_data}節のスナップ工程で得られる
% 力・トルク成分の波形データの内，各力・トルク成分それぞれをデータセットとして関数主成分分析を適用する．
% これにより，該当する力・トルク成分の波形を最も良く表現する複数の主成分関数が得られる．
% その後，データセット内の各波形が，各主成分関数をどの程度多く含んでいるのかを個別に評価する指標値として関数主成分得点を算出し，これを特徴量ベクトルとして用いる．
% 以上の行程を力・トルクデータ全てに対して適用することで，一つのオフセットパターンに対して，各力・トルク成分に対応する6つの特徴量ベクトルを抽出する．
% ここで，抽出した6つの特徴量ベクトルを1つの特徴量ベクトルとしてまとめて以降の学習を行う場合，これら特徴量ベクトルは各力・トルク成分波形データから独立に取得されるため，特定の力・トルク成分に大きく依存する特徴量ベクトルが取得され，成否パターンの分類が困難となることが予想される．
% よって本研究では，各力・トルク成分から得られたそれぞれの特徴量ベクトルに対して，個別にSVMによる学習を実行する．
% 右探り動作工程，左探り動作工程についても，上記と同様にして特徴量抽出を行う．

\subsubsection{Decision tree based on SVM} 
\label{3.4}

\begin{figure}[t]
% \begin{minipage}[b]{0.49\linewidth}
%  \centering
%  \includegraphics[width=0.5\linewidth]{picture/chap3/dy_5mm_case.eps}
%  %\includegraphics[height=5cm,bb =0 0 427 609]{picture/chap3/dy_5mm_case.eps}
%  \subcaption{Assembly simulation with $\Delta y = 5[{\mathrm{mm}}]$,$\Delta \theta_z = 0[^\circ]$}
% \end{minipage}
% \begin{minipage}[b]{0.49\linewidth}
%  \centering
%  \includegraphics[width=0.5\linewidth]{picture/chap3/y-0.5_image.eps}
%  %\includegraphics[height=5cm,bb =0 0 427 609]{picture/chap3/y-0.5_image.eps}
%  \subcaption{Assembly simulation with $\Delta y = -5[{\mathrm{mm}}]$,$\Delta \theta_z = 0[^\circ]$}
% \end{minipage}\\\\
 \begin{minipage}[b]{0.49\linewidth}
  \centering
  \includegraphics[width=\linewidth]{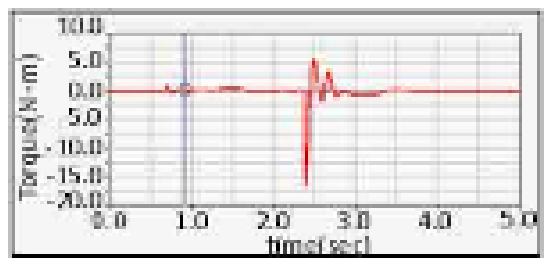}
  \subcaption{Torque about $y$ axis:1}
 \end{minipage}
 \begin{minipage}[b]{0.49\linewidth}
  \centering
  \includegraphics[width=\linewidth]{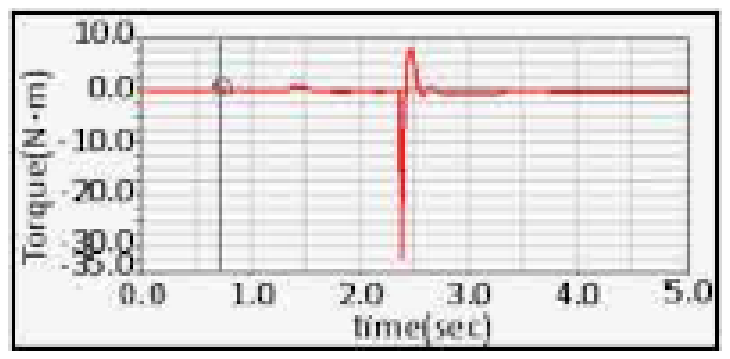}
  \subcaption{Torque about $y$ axis:2}
 \end{minipage}
 \caption{Torque about $y$ axis obtained through experiments with two different error states}\label{fig:problem_of_clustering}
\end{figure}

\begin{figure}[t]
 \begin{minipage}[b]{0.49\linewidth}
  \centering
  \includegraphics[width=\linewidth]{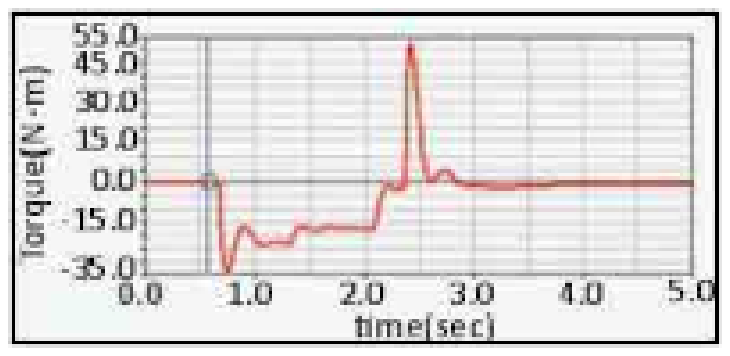}
  \subcaption{Torque about $x$ axis:1}
 \end{minipage}
 \begin{minipage}[b]{0.49\linewidth}
  \centering
  \includegraphics[width=\linewidth]{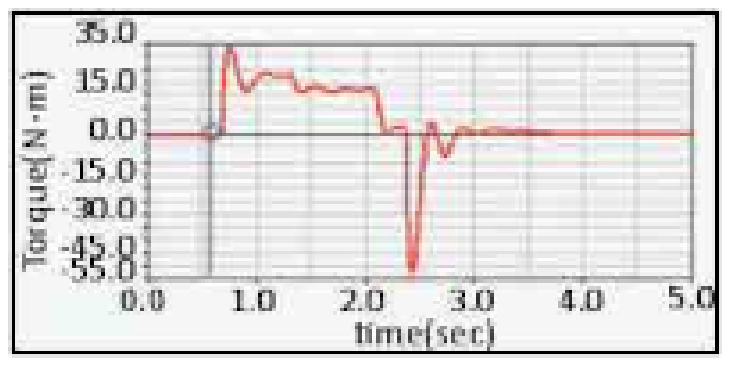}
  \subcaption{Torque about $x$ axis:2}
 \end{minipage}
 \caption{Torque about $x$ axis obtained through experiments with two different error states}\label{fig:hint_of_clustering}
\end{figure}

In this subsection, we elucidate the construction of the decision tree for classifying error states. 
Here, depending on the error state, the component of force/torque that can enable the classification of the error state is different. 
For example, let us consider the force/torque profiles corresponding to two assembly tasks with different offset patterns shown in Figs. \ref{fig:problem_of_clustering} and \ref{fig:hint_of_clustering}. 
If we compare the torques about the $y$-axis, it becomes extremely difficult to differentiate between two offset patterns, as shown in Fig. \ref{fig:problem_of_clustering}. However, if we compare the torques about the $x$-axis, we can easily differentiate between two offset patterns, as shown in Fig. \ref{fig:hint_of_clustering}. 

% 本節では，\ref{3.3_sec:feature_extraction}節により抽出された特徴量ベクトルが
% 持つ成否パターンを分類する決定木を構築する．

% ここで，取得される力・トルクの波形データはスナップアセンブリ時の作業部品同士の物理作用を反映したものであるため，得られる特徴量については，分類可能な成否パターンが各力・トルク成分ごとに異なると考えられる．
% ここでは，波形の特徴がより明確に表れる物理シミュレータ"adams"\cite{adams}による
% スナップアセンブリの物理シミュレーションを用いて原理の説明を行う．
% 図\ref{fig:problem_of_clustering}(a)，(b)のような成否パターンの異なる2つのスナップアセンブリで取得される$y$軸まわりのトルクの波形データである図\ref{fig:problem_of_clustering}(c)，(d)を比較したところ，同様の波形データが確認された．

\fboxsep=0pt 

% このような場合，波形データから取得される特徴量も成否パターンが反映されていないものであるため，異なる成否パターンに対して近い値の特徴量が取得されることが予想される．よって，$y$軸まわりのトルクから取得される特徴量からは図\ref{fig:problem_of_clustering}(a)，(b)の成否パターンの分類が不可能であることが考えられる．
% 一方，図\ref{fig:problem_of_clustering}(a)，(b)の成否パターンにおける$x$軸まわりのトルクの波形データ図\ref{fig:hint_of_clustering}(a)，(b)を比較したところ，波形に顕著な反転性が確認された．このような波形であれば，波形データから抽出される特徴量は成否パターンが反映されており，図\ref{fig:problem_of_clustering}(a)，(b)の成否パターンが分類可能であると考えられる．

Based on this observation, we construct a decision tree wherein we 
divide the training data included in the each node into two groups by referring to an appropriate force/torque component. 
The algorithm employed for constructing the decision tree is described in Algorithm \ref{algorithm_SVM}. 

% このことから，力・トルク成分の特徴量から，部分的に成否パターンを分類できる特定の力・トルク成分から順番に分類を行い，分類結果に基づいて特徴量の分離，除外を繰り返すことで詳細な成否パターンの分類が可能であると考えられる．本研究では，分類にSVMを用いて，\ref{3.1}節で教師データとして定義された成否パターン(1)$\sim$(9)の決定木を構築することで上記を達成する．以下，\ref{3.3_sec:feature_extraction}節で取得される各力・トルク成分の特徴量分布に対して，最良の決定木が得られるような，力・トルク成分に対するSVMの実行順番，及びその際分類する成否パターンの組を決定するアルゴリズム\ref{algorithm_SVM}を示す．
\begin{algorithm}
  \caption{Classifier constructed based on SVM}  \label{algorithm_SVM}
  \SetKwData{Null}{null}
  \SetKwFunction{append}{append}
  \SetKwFunction{lowest}{lowest}
  \SetKwFunction{isAvailable}{isAvailable}
  \SetKwFunction{getHeight}{getHeight}
  \SetKwFunction{getCurves}{getCurves}
  \SetKwFunction{addToSet}{addToSet}
  %\DontPrintSemicolon
  \KwData{fPCA scores and success/error states \newline
               Success pattern index: $i \leftarrow 0$\newline
               Error pattern index: $i \leftarrow 1,\cdots,F$\newline
               Force/torque component: $j \leftarrow 1,\cdots,6$\newline
               Index of waveform data: $k \leftarrow 1,\cdots, N(i)$\newline
               Waveform data: ${\rm WD}(i,j,k)$\newline
               Combinations of success/error states: $C$
               \\}
  \KwResult{Construction of a decision tree }
% \end{algorithm}
% \begin{algorithm}
  \Begin {
            ${\rm id} \leftarrow 0$\\
            ${\rm node}({\rm id}){\rm .pattern\_id} \leftarrow [0,1,\cdots,F]$\\
            ${\rm max\_id} \leftarrow 0$\\
            
                   \While{1}{
                      \If{${\rm size}({\rm node}({\rm id}).{\rm pattern\_id}==1)$}{
                          \If{${\rm id}=={\rm max\_id}$}{
                              return node\\
                          }
                          ${\rm id} \leftarrow {\rm id}+1$\\
                          continue\\
                      }
                      \For{$j \leftarrow 1, \cdots, 6$}{
                         \For{$^{\forall}{C} \subset {\rm node}({\rm id}){\rm .pattern\_id}$}{
                            $i\leftarrow {\rm node}({\rm id}){\rm .pattern\_id}$,\\
                            ${\rm patterns}(j,C) \leftarrow$ SVM(${\rm fPCAScore}({\rm WD}(i,j,k)),C$)\\
                            ${\rm accuracy}(j,C) \leftarrow {\rm calcAccuracy}({\rm patterns}(j,C))$\\
                         }
                     }
                     ${\rm node}({\rm id}).{\rm component} \leftarrow {\rm argmax_{j}}({\rm accuracy}(j,C))$\\
                     ${\rm node}({\rm id}){\rm .patterns} \leftarrow {\rm argmax_{C}}({\rm accuracy}(j,C))$\\
                     ${\rm node}({\rm id}).{\rm children} \leftarrow [{\rm max\_id}+1, {\rm max\_id}+2]$\\
                    
                     ${\rm node}({\rm max\_id}+1){\rm .pattern\_id} \leftarrow {\rm node}({\rm id}){\rm .patterns}$\\
                     ${\rm node}({\rm max\_id}+2){\rm .pattern\_id} \leftarrow {\rm node}({\rm id}){\rm .pattern\_id}-{\rm node}({\rm id}){\rm .patterns}$\\
                     ${\rm max\_id} \leftarrow {\rm max\_id}+2$\\
                     ${\rm id} \leftarrow {\rm id}+1$\\
                   }
                   %${\rm return}\ {\rm node}$\\
  }
\end{algorithm}

\noindent
In the initial state, the decision tree is composed only of the root node that includes all the training data. In this node, we iteratively apply the training data corresponding to each force/torque component to the SVM to split the training data into two groups such that the highest accuracy is obtained. If we can find such force/torque component, we add two children nodes and split the training data into two groups. We iterate these steps until we can classify each error state.  We calculate the accuracy of classification based on the following equation:
%
% アルゴリズム\ref{algorithm_SVM}について説明する．まず，初期状態では決定木は根ノードのみを保持しており，全オフセットパターンそれぞれに対する，各力・トルク成分における6つの特徴量ベクトル，及び教師データとして与えた成否パターン(1)$\sim$(9)のいずれかとの組が格納されている．
% このノードにおいて，ノードの持つ成否パターンの全通りの組み合わせで選択された成否パターン群それぞれに対して，選択されなかった成否パターンとの分類を行うようにSVMを繰り返し実行し，各分類結果に対して分類精度を求める．これを各力・トルク成分の持つ特徴量分布それぞれにおいて実行する．
% そして，最も分類精度の高い力・トルク成分と分類される成否パターンの組を求め，左右の子ノードに分類された成否パターンと，対応する特徴量ベクトルを格納する．
% 提案手法では，以下の式によってSVMの分類精度を計算する．
\begin{eqnarray}
Accuracy = \frac{\frac{\sum P_{TP}+\sum P_{FP}}{TP+FN}+\frac{\sum P_{FN}+\sum P_{TN}}{FP+TN}}{2},
\label{accuracy}
\end{eqnarray}
% ここで，$TP$(True Positive)，$FP$(False Positive)はそれぞれ分類された2群の成否パターンに対して真値と同じ分類結果であった特徴量の数であり，$TN$(True Negative)，$FN$(False Negative)は真値と異なる分類結果であった特徴量の数である．
where 
$TP, FP, TN$ and $FN$ denote the true positives, false positives, true negatives and false negatives, respectively, corresponding to the feature quantities, and $P_*$ denotes the class probability of $*$ based on the output of SVM. 

%↓再現率や得意率は修論時点の評価関数
%つまり，
% $\frac{TP}{TP+FN}$は再現率，$\frac{TN}{FP+TN}$は特異率である．
% %\end{center}

% 以上の手順を全ての葉ノードにおいて，格納されている成否パターンがただ1つになるまで繰り返すことで，成否パターンを完全に分類する決定木が構築される．

% 以上の処理をスナップ工程，右探り動作工程，左探り動作工程それぞれ個別に行うことで，3種類の決定木を構築する．

\subsection{Identification of error states and error recovery} 
\label{3.5}

In this subsection, we present the identification of an error state before the error actually occurs, after which a robot tries to recover from the error state. 
The identification of error states in this study includes two steps. 
In the first step, the male part is moved in the $+z$ direction so that it can be fit to the female part; this will hereafter be called the assembly motion. 
Before the error actually occurs, the robot stops moving the male part and attempts to identify an error state. 
However, if the class probability is not high enough, we move on to the second step. 
In the second step, we perform additional probing to identify the error state with a higher probability. 

% 本節では，学習結果をもとに作業途中での成否パターンの識別を行い，その後，識別された成否パターンに基づくエラーリカバリ動作の生成を行う．

% はじめに，作業途中での成否パターンの識別方法について説明する．
% 提案手法では，スナップ工程と探り動作工程の二段階に分けて識別を行い，スナップ工程のみでは正しい識別が困難である場合に限り，左右の探り動作工程に移行することで再度識別を行う

During the assembly motion, we obtain the 6D force/torque profile, which is 
terminated before the error actually occurs. Using this profile, we obtain the feature vectors 
as described in subsection \ref{3.3_sec:feature_extraction}. The feature vector 
is applied to the decision tree elucidated in subsection \ref{3.4}. 
We check the class probability of all the nodes employed to identify the error state. 
If the class probability of at least one of the nodes is lower than the threshold, it is difficult to correctly estimate the error state. 
%%分類スコアと分類精度の違いは何か分からなかった。。。%%%%%%%%
In this case, the male part entails additional probing to identify the error sate more correctly. It first moves in the $+x$ direction and then in the $-x$ direction. 
We note that we have also constructed decision trees for additional probing in $+x$ and $-x$ directions in the offline phase.  
We obtain the 6D force/torque profiles for both cases, and then apply the feature vector to the decision tree to more accurately identify the error state. We apply the identification result with the higher accuracy, calculated using eq.(\ref{accuracy}) of the node with the smallest accuracy.  

% まず，未知のオフセットパターン下にてスナップ工程を行い，その間に取得される力・トルクデータから
% \ref{3.3_sec:feature_extraction}節により特徴量を抽出する．そして，\ref{3.4}節で構築されるスナップ工程における決定木を根ノードから参照して葉ノードまで到達させ，経由した各ノードにおいて，
% 学習データと未知の特徴量との類似度を示す値である分類スコアを保持する．
% そして，以下の処理を行うことで探り動作の必要性を判定し，成否パターンの識別を行う．

% \begin{itemize}
% \item 各ノードでの分類スコアの絶対値に対して，あらかじめ設定した閾値以下のノードが存在せず，かつ，経由ノードの中に閾値以下の分類精度値が存在しない場合
% 場合\newline
% 経由ノードでの特徴量が，SVMによる分離境界外に存在しており，識別結果の信頼度が高く，スナップ工程のみで正しい識別が達成されていると想定される．
% さらに，分類精度値が高いことから，適切なノード生成が成されていると考えられる．
% よって，葉ノードが持つ成否パターンを識別結果とする．
% \item 各ノードでの分類スコアの絶対値に対して，あらかじめ設定した閾値以下のノードが存在する場合
% ，または，経由ノードの中に分類精度値が閾値のノードが存在する場合
% \newline
% 前者の場合，該当ノードでの特徴量が，SVMによる分離境界付近に存在しており，識別結果の信頼度が低く，スナップ工程のみでは正しい識別が困難であると想定される．
% また，後者の場合，該当ノードの分類精度が低く，抽出される特徴量にかかわらず不適切な識別が成される恐れがある．
% 以上の場合，探り動作工程に移行する．左右の探り動作時に取得した力・トルク情報に対して，上記と同様にして左右の探り動作それぞれについて経由ノードでの分類スコアを保持する．そして，左探り動作工程と右探り動作工程の持つ分類スコアの内，絶対値が最小のものを比較する．そして，より大きい絶対値を持つ方向への探り動作をより信頼度の高い探り動作であるとして，該当する探り動作工程における葉ノードが持つ成否パターンを識別結果とする．
% \end{itemize}

Finally, in this section, the method for error recovery is explained as follows: 
If the assembly is predicted to be successful, the male part is further moved in the $+z$ direction and fit to the female part. 
If an error state is identified, the male part is moved once in the $-z$ direction and then moved by a fixed (small) distance in the direction opposite to that in the identified error state. Then, the male part is moved in the $+z$ direction again to fit it to the female part. 

% 次に，識別後のエラーリカバリ動作について説明する．
% 上記の識別結果が成功パターンであるかどうかによって，以下の処理を行うことでエラーリカバリシステムを実現する．

% \begin{itemize}
% \item 成功パターンと識別された場合\newline
% エラーリカバリ動作は不要である．よって，ロボットハンドを$+z$方向に数ミリ下ろすことでスナップアセンブリを完了する．
% \item エラー状態(2)$\sim$(9)と識別された場合\newline
% 識別された成否パターンの持つオフセットを解消する方向に数ミリ，または数度ロボットハンドを移動させることでエラー状態からのリカバリを行う．また同時に，ロボットハンドを$-z$方向に移動させることで，スナップアセンブリ作業の初期状態に戻り，スナップアセンブリ作業へ復帰する．
% \end{itemize}

\color{black}

\section{Experiment}
\label{4}

This section detailss the experiment conducted to demonstrate the effectiveness of the proposed approach. 

%本章では，提案手法の有用性を検証するために行った実験について説明する．
\subsection{Experimental environment}
\label{4.1}

\begin{figure}[t]
  \begin{center}
          \includegraphics[width=0.9\linewidth]{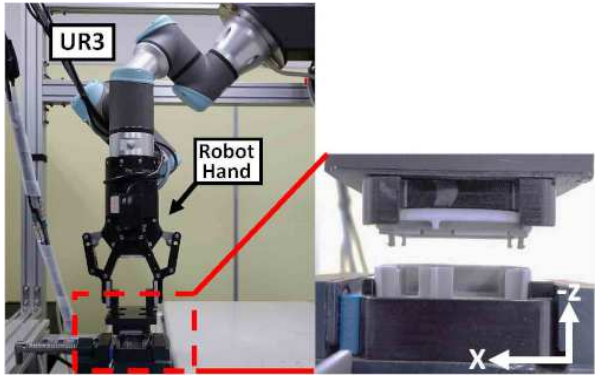}
    \caption{Experimental environment}
    \label{experimental_environment}
  \end{center}
\end{figure}

\begin{figure}[t]
 \begin{minipage}[b]{0.49\hsize}
  \centering
  \includegraphics[width=\linewidth]{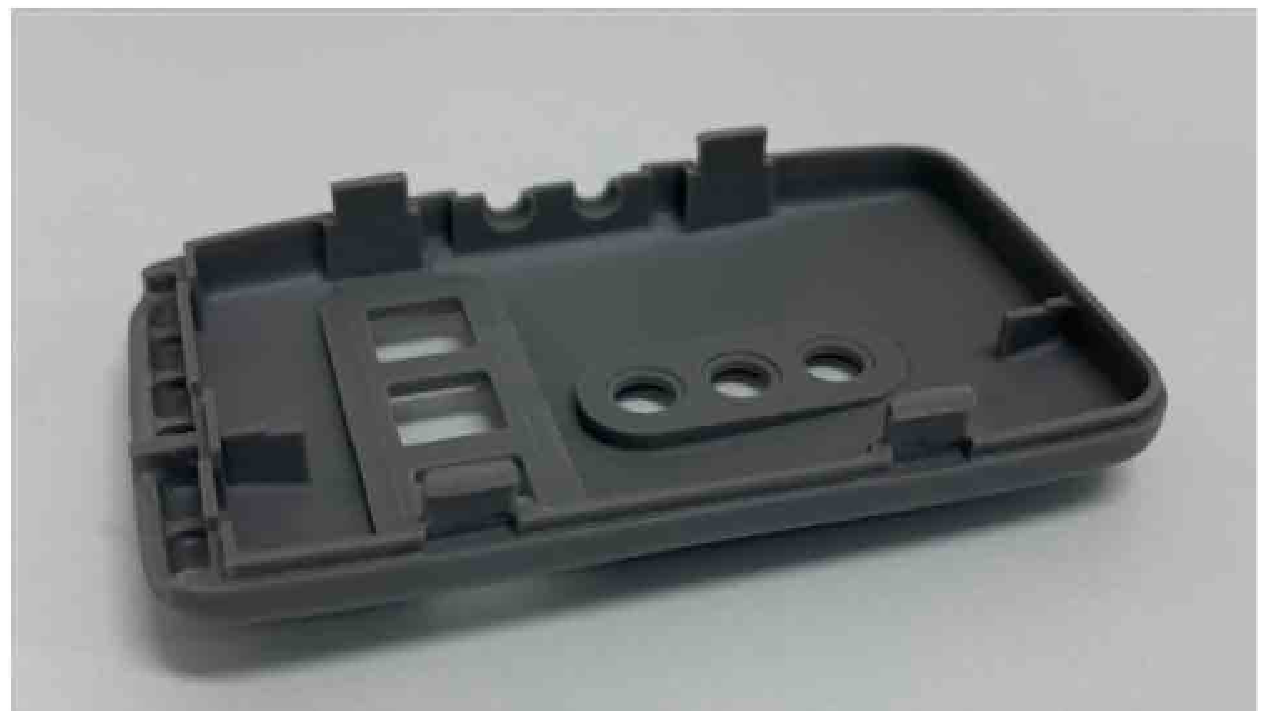}
  \subcaption{Male part}
 \end{minipage}
 \begin{minipage}[b]{0.49\hsize}
  \centering
  \includegraphics[width=0.82\linewidth]{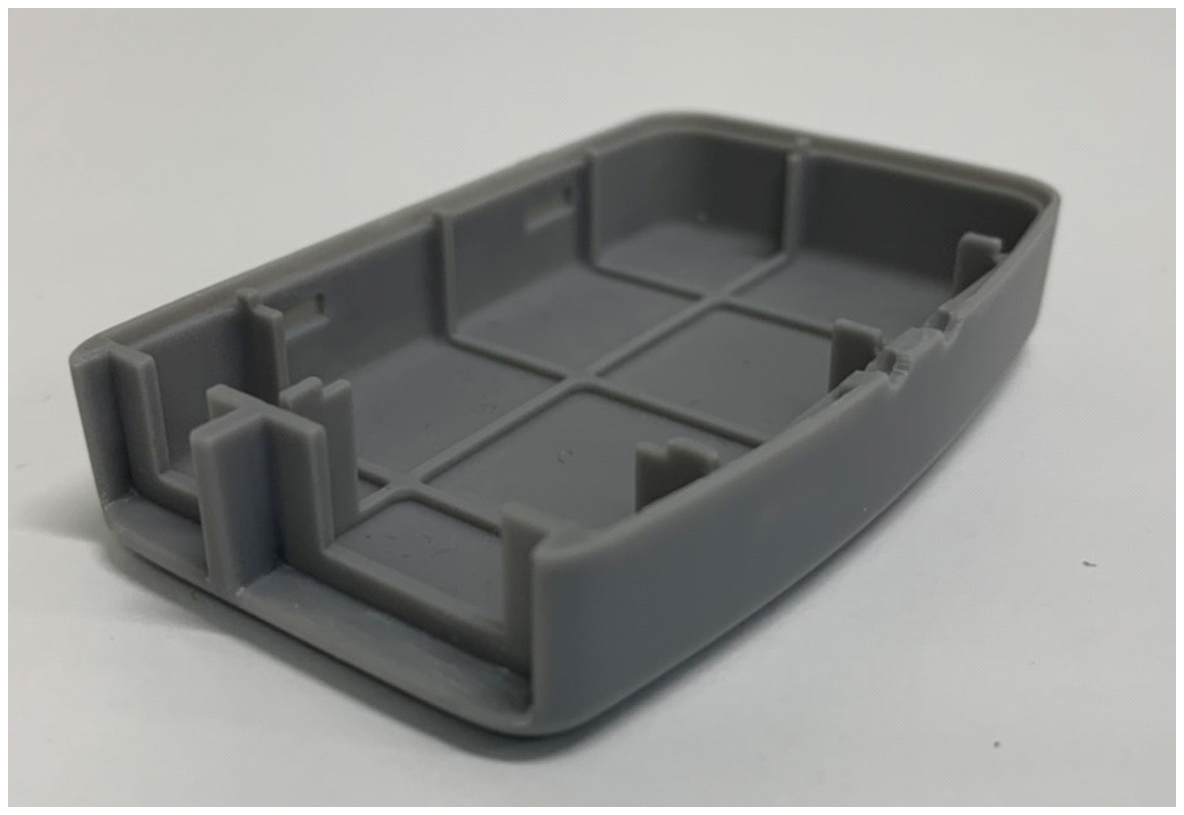}
  \subcaption{Female part}
 \end{minipage}\\
 \caption{Assembly parts}
    \label{parts}
\end{figure}

Our experimental setup is shown in Fig. \ref{experimental_environment}. 
We used UR3 with a Robotiq 6D force/torque sensor attached at the wrist. 
To avoid the application of an excessively large force at the wrist, we also installed a compliant mechanism at the wrist (SHM61J, Koganei Co., Ltd.). 
The plastic parts with four snap joints used for assembly are shown in Fig. \ref{parts}. 

% 本実験を行う際の実験環境
% を図\ref{experimental_environment}に示す．
% 作業ロボットとして，
% Universal Robot社のUR3\cite{UR3}を使用した(図\ref{experimental_environment})．
% また，グリッパはRobotiq社の"Robotiq85"\cite{robotiq85}を使用した．
% UR3のロボットハンド部の概観を図\ref{hand}に示す．
% ここで，スナップアセンブリ時にロボットハンドに過大な力が加わることを避けるために，図\ref{hand}-(1)のようにロボットハンドの手首部分にコンプライアンス機構\cite{compliance unit}を設ける．

%\begin{figure}[!h]
%  \begin{center}
%          \includegraphics[width=0.5\linewidth]{picture/chap4/setting/hand.eps}
%    \caption{Robot hand}
%    \label{hand}
%  \end{center}
%\end{figure}

%次に，実験で用いた組み立て対象部品を図\ref{parts}に示す．

% 図\ref{parts}(a)の突起を有する部品をロボットハンドで把持し，図\ref{parts}(b)の部品に嵌め込むことでスナップ工程，及び探り動作工程が実行される．そしてその間，図\ref{hand}-(2)の6次元力センサによって力・トルク波形データを取得する．本研究では力センサとしてRobotiq社の"FT300 Force Torque Sensor"\cite{FT sensor}を用いる．
% ここで，図\ref{parts}(a)の部品を図\ref{hand}-(3)のグリッパ部で直接把持する場合，嵌め合い時に発生する力の影響により部品がロボットハンドの把持位置からずれることで，意図しないオフセット値が加わり，不適切な波形データが取得される恐れがある．そこで，図\ref{grasp helper}(a)に示す補助パーツを突起部品に固定させ，同様に補助パーツをグリッパに固定させることで上記を解決する．突起を有する部品に補助パーツを取り付けた部品を図\ref{grasp helper}(b)に示す．

%\begin{figure}[!h]
% \begin{minipage}[b]{0.49\linewidth}
%  \centering
%  \includegraphics[width=0.8\linewidth]{picture/chap4/setting/grasp_helper.eps}
%  \subcaption{Auxiliary part}
% \end{minipage}
% \begin{minipage}[b]{0.49\linewidth}
%  \centering
%  \includegraphics[width=0.8\linewidth]{picture/chap4/setting/helper_male_part.eps}
%  \subcaption{Combined part}
% \end{minipage}\\
% \caption{Auxiliary part and combined part}
%    \label{grasp helper}
%\end{figure}

\subsection{Force/torque data}
\label{4.2}

We performed assembly experiment with 131 offset patterns assuming different initial position/orientation of the male part in the range of 
$-2.0[{\mathrm{mm}}]\leq \Delta x\leq 2.0[{\mathrm{mm}}]$, $-2.0[{\rm deg}]\leq \Delta \theta_z\leq 2.0[{\rm deg}]$. 
In an assembly task, the male part is first moved 6$[\mathrm{mm}]$ to the $+z$ direction, and the error states are identified. If additional probing is required, the male part is first moved 1$[\mathrm{mm}]$ to the $-z$ direction and then $2[\mathrm{mm}]$ to the $\pm{x}$ directions. 
The threshold of the class probability introduced in subsection \ref{3.5} is set at $0.2$. To recover from an error state, the male part is moved $1[\mathrm{mm}]$ and 1[deg] in the opposite direction of the identified error state. 

% 次に，発生させるオフセット値の範囲については，製造業の現場では，治具が経年劣化した場合，発生するオフセット値は
% $-2.0[{\mathrm{mm}}]\leq \Delta x\leq 2.0[{\mathrm{mm}}]$，$-2.0[^\circ]\leq \Delta \theta_z\leq 2.0[^\circ]$程度である．
% そこで，本実験においても同様の範囲において，$0.5[\mathrm{mm}]，[^\circ]$間隔でオフセットを発生させ，計$81$種程度のオフセットパターンのもと波形データを取得する．
% また，本実験環境では作業部品を固定する治具が経年劣化した状態を再現することが困難であるため，嵌め合い対象部品に対して\ref{3}章で述べたオフセットを発生させることが不可能である．そこで，万力で嵌め合い対象部品を固定させ，ロボットハンドをあらかじめ初期状態から$\pm{x}$，及び$\pm{\theta_z}$方向に移動させた状態からロボットハンドを下ろして嵌め合いを行うことで作業部品間のオフセットを再現し，スナップアセンブリを実行する．

% また，スナップアセンブリの各工程でのロボットハンドの移動量について，
% スナップ工程では対象部品の嵌め合いが完了するよりも前の時点で工程を完了させる必要がある．
% また，探り動作工程では，部品の突起部分が破損せず，かつ全体の作業時間に影響を及ぼさない程度の移動量にすることが望ましい．
% そこで，上記について予備実験を行った．
% 予備実験の結果，嵌め合い対象部品の鉛直真上に位置する既知の初期状態からスナップアセンブリを開始し，ロボットハンドに対してオフセットを発生させた後，
% スナップ工程として，$6[\mathrm{mm}]$下($+z$)方向に移動させる．その後，探り動作が必要であれば，$1[\mathrm{mm}]$ロボットハンドを上($-z$)方向に上げ，その後，$\pm{x}$方向に$2[\mathrm{mm}]$ロボットハンドを往復させることで探り動作工程を行うこととした．
% 以上の条件を学習段階，識別段階の両方で統一させて実験を行った．

% 最後に，識別を行う際，\ref{3.5}節の閾値については，
% 分類スコア，分類精度値ともに$0.2$とし，
% エラーリカバリ動作時のロボットハンドの移動量については，本実験ではオフセットを解消させる方向へ$1[\mathrm{mm}]，[^\circ]$移動させることとする．

\begin{figure}[t]
 \begin{minipage}[b]{0.49\linewidth}
  \centering
  \includegraphics[width=\linewidth]{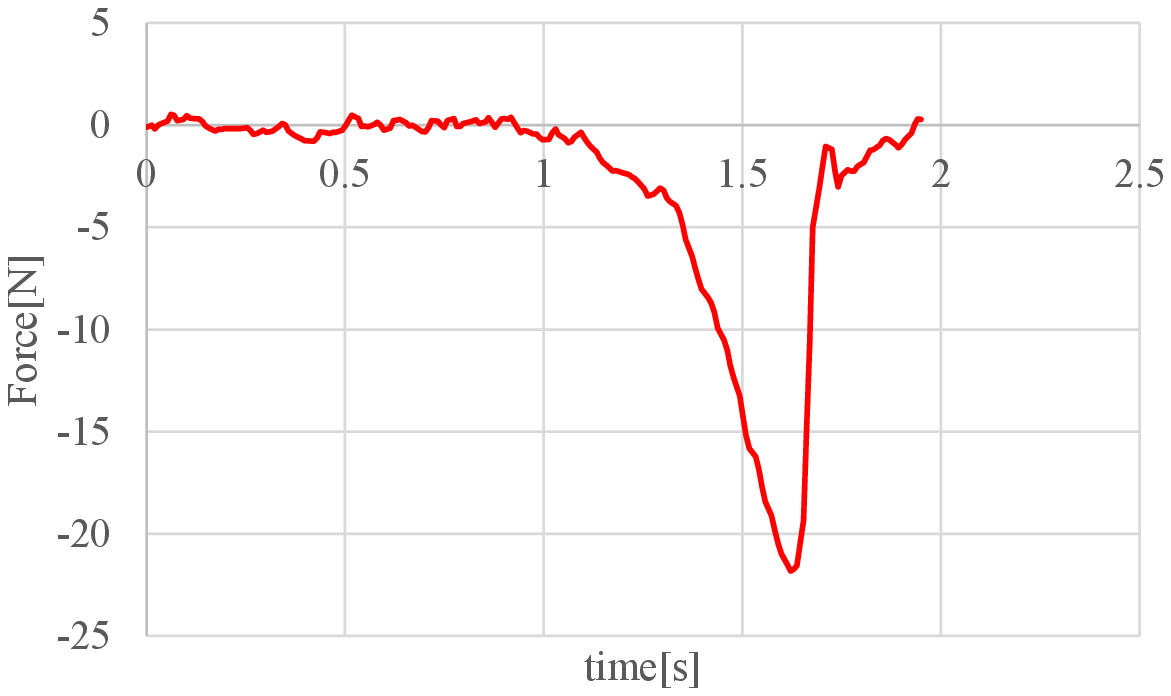}
  \subcaption{Successful case (1)}
 \end{minipage}
 \begin{minipage}[b]{0.49\linewidth}
  \centering
  \includegraphics[width=\linewidth]{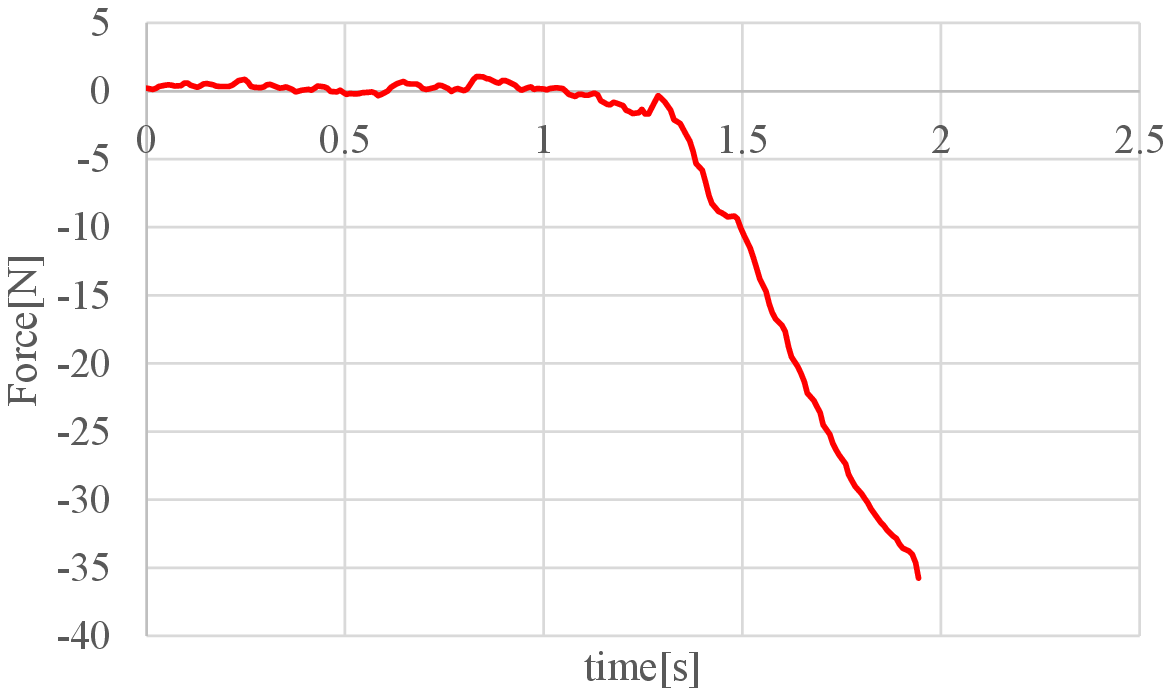}
  \subcaption{Error state (6)}
 \end{minipage}
 \caption{Force in $z$ direction obtained through experiment of successful and failure cases}
    \label{waveform Fz}
\end{figure}

\begin{figure}[t]
 \begin{minipage}[b]{0.49\linewidth}
  \centering
  \includegraphics[width=\linewidth]{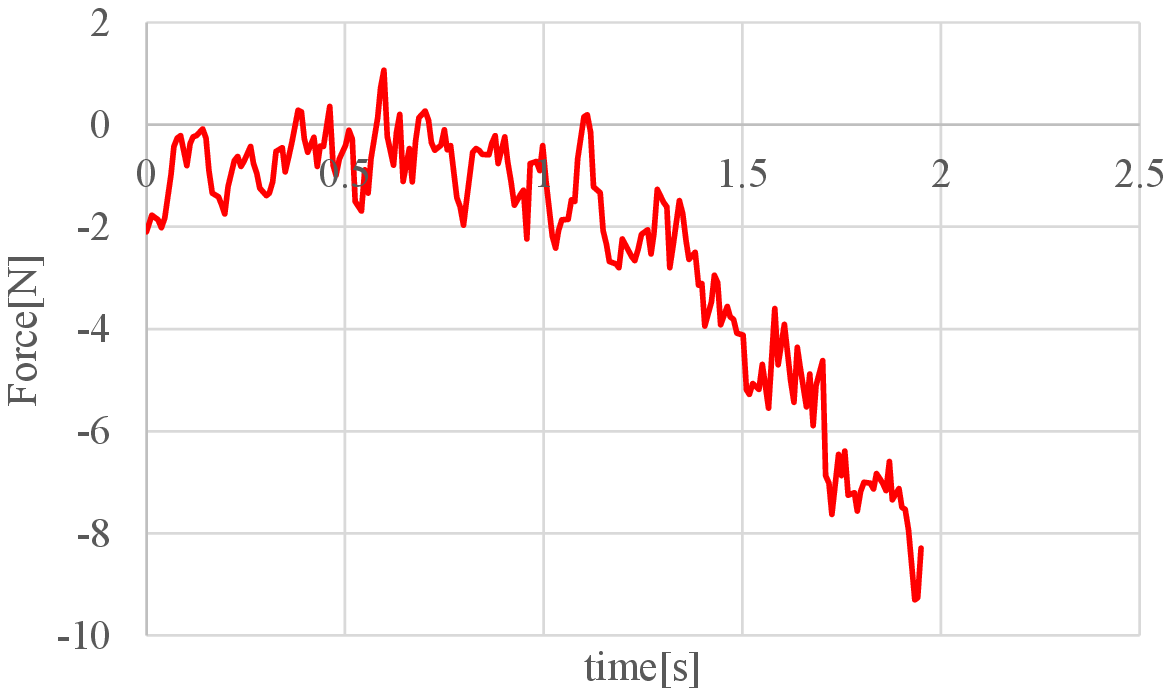}
  \subcaption{Error state (7)}
 \end{minipage}
 \begin{minipage}[b]{0.49\linewidth}
  \centering
  \includegraphics[width=\linewidth]{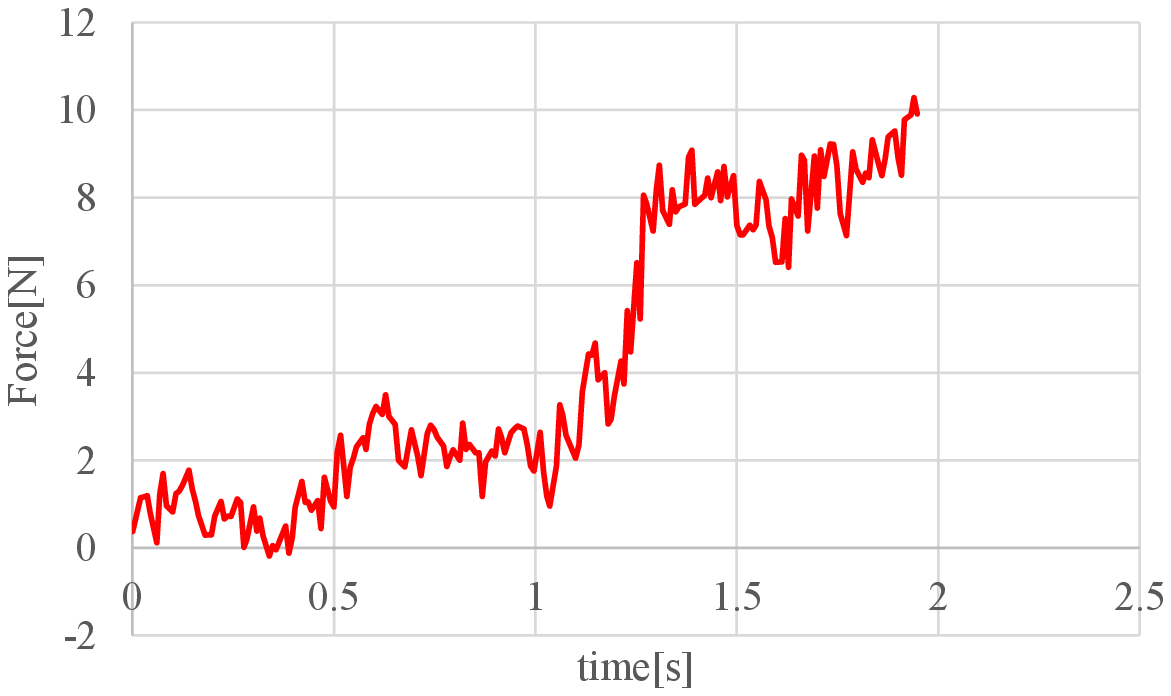}
  \subcaption{Error state (8)}
 \end{minipage}
 \caption{Force in $x$ direction obtained through experiment of different error states}
    \label{waveform Fx}
\end{figure}

\begin{figure}[t]
 \begin{minipage}[b]{0.49\linewidth}
  \centering
  \includegraphics[width=\linewidth]{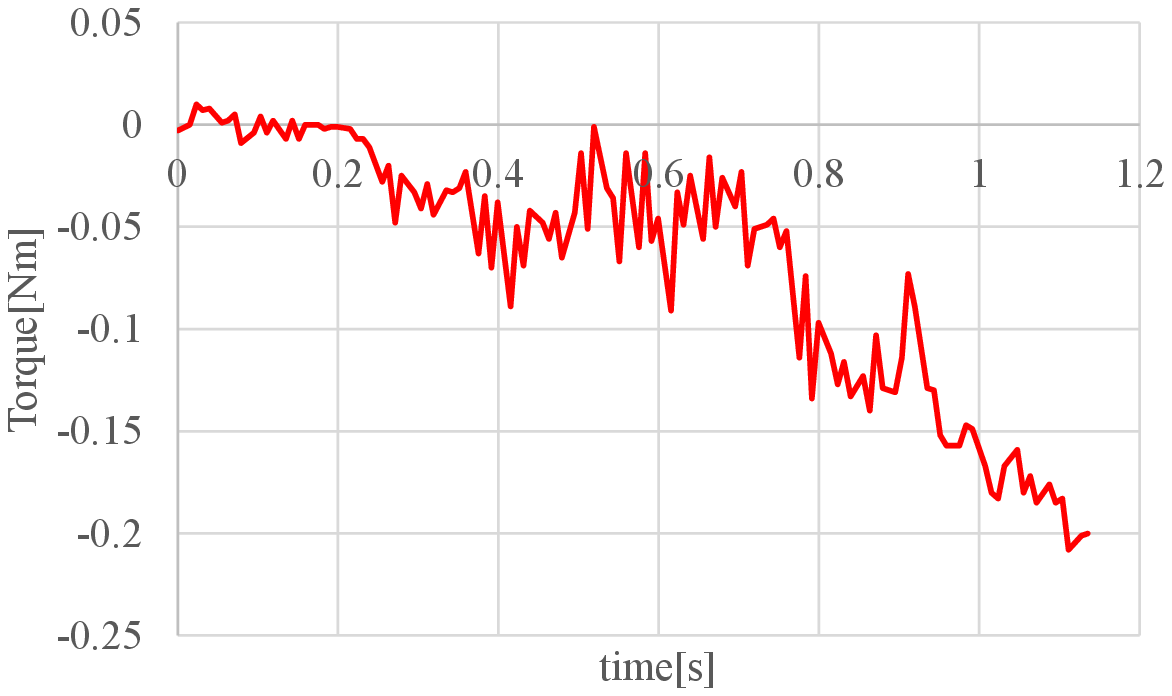}
  \subcaption{Error state (2)}
 \end{minipage}
 \begin{minipage}[b]{0.49\linewidth}
  \centering
  \includegraphics[width=\linewidth]{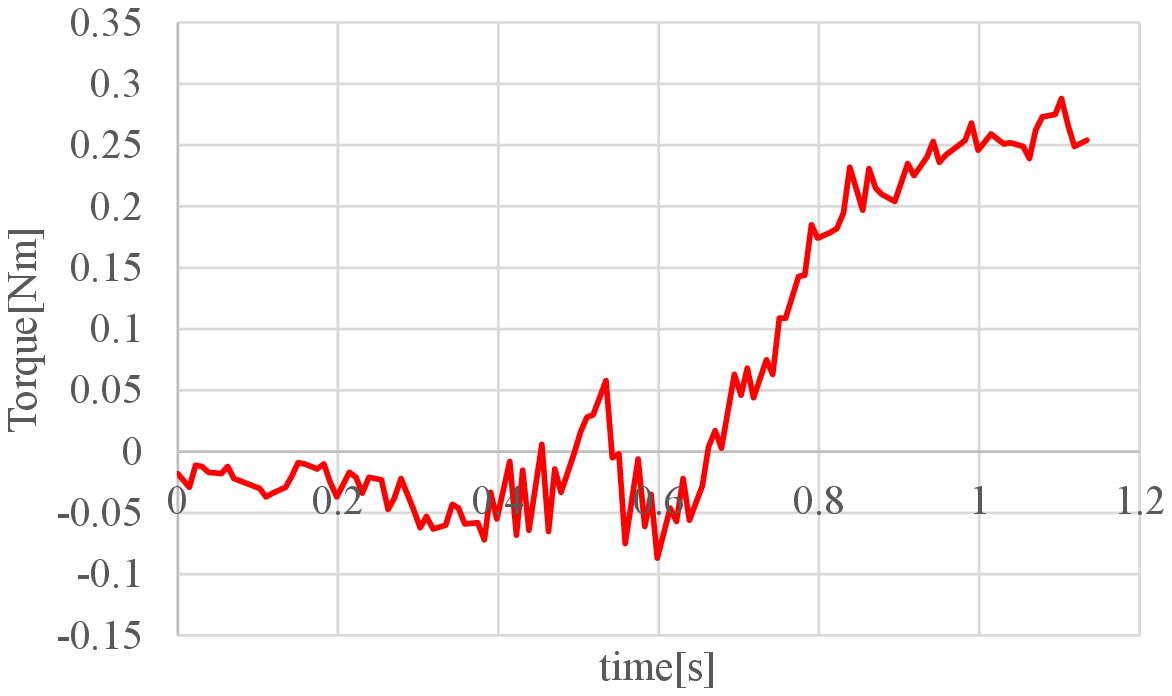}
  \subcaption{Error state (3)}
 \end{minipage}
 \caption{Torque about $x$ axis obtained through experiment of different error states with additional probing in the $+x$ direction}
    \label{waveform Tx}
\end{figure}

\begin{figure}[t]
 \begin{minipage}[b]{0.49\linewidth}
  \centering
  \includegraphics[width=\linewidth]{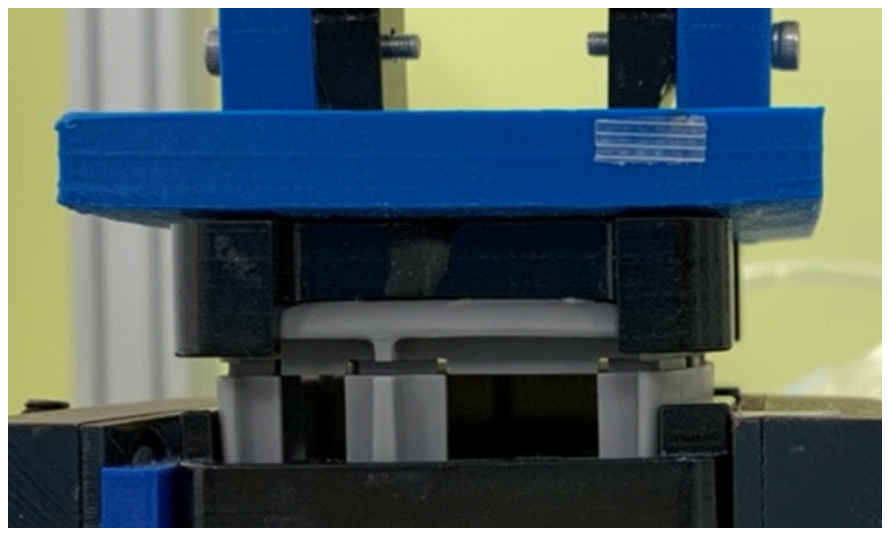}
  \subcaption{Front view of assembly}
 \end{minipage}
 \begin{minipage}[b]{0.49\linewidth}
  \centering
  \includegraphics[width=0.86\linewidth]{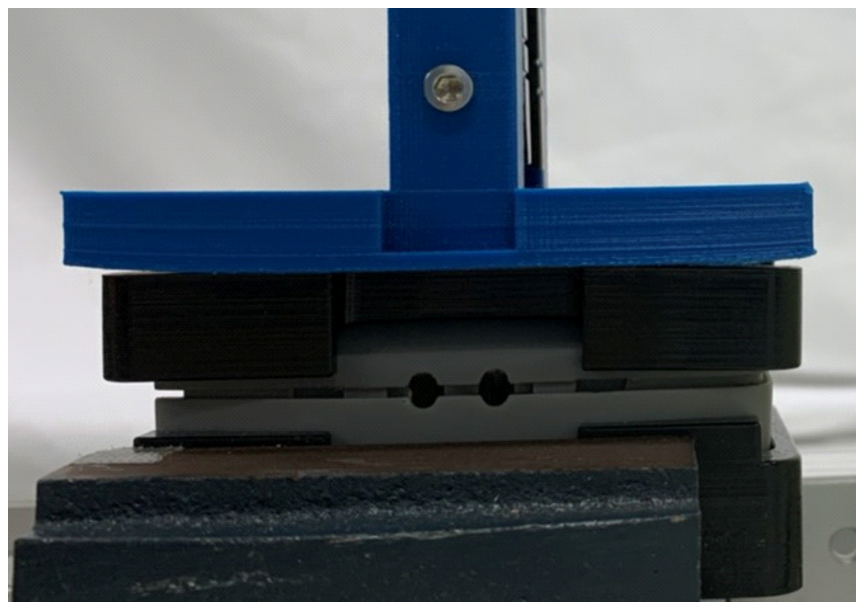}
  \subcaption{Side view of assembly}
 \end{minipage}\\
 \begin{minipage}[b]{0.49\linewidth}
  \centering
  \includegraphics[width=0.95\linewidth]{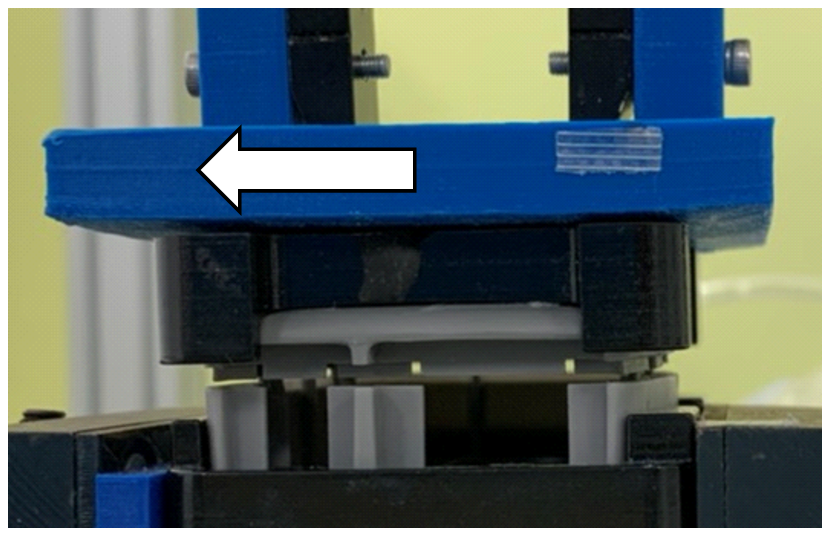}
  \subcaption{Additional probing in $+x$ direction}
 \end{minipage}
 \begin{minipage}[b]{0.49\linewidth}
  \centering
  \includegraphics[width=\linewidth]{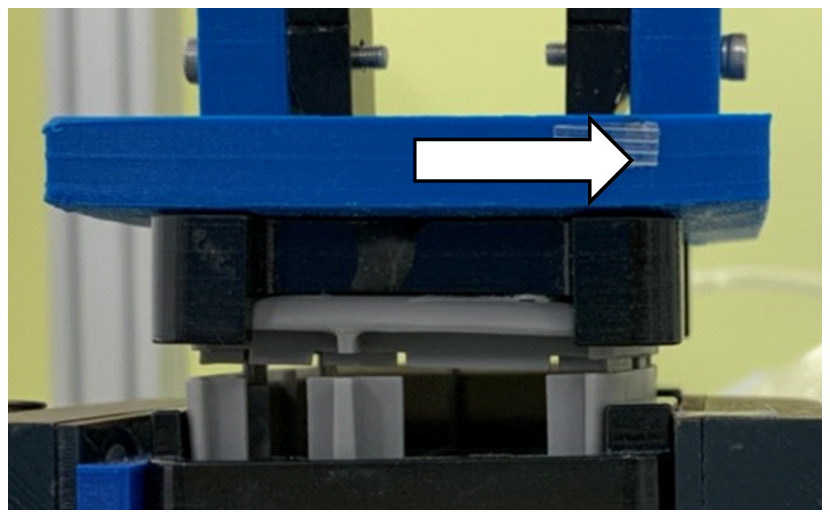}
  \subcaption{Additional probing in $-x$ direction}
 \end{minipage}
 \caption{Snapshots of successful snap assembly ($\Delta y=0[\mathrm{mm}]$, $\Delta \theta_z=$0.5[deg]) }
     \label{snapshots success}
\end{figure}

\begin{figure}[t]
{\centering
 \begin{minipage}[b]{0.4\linewidth}
  \centering
  \includegraphics[width=\linewidth]{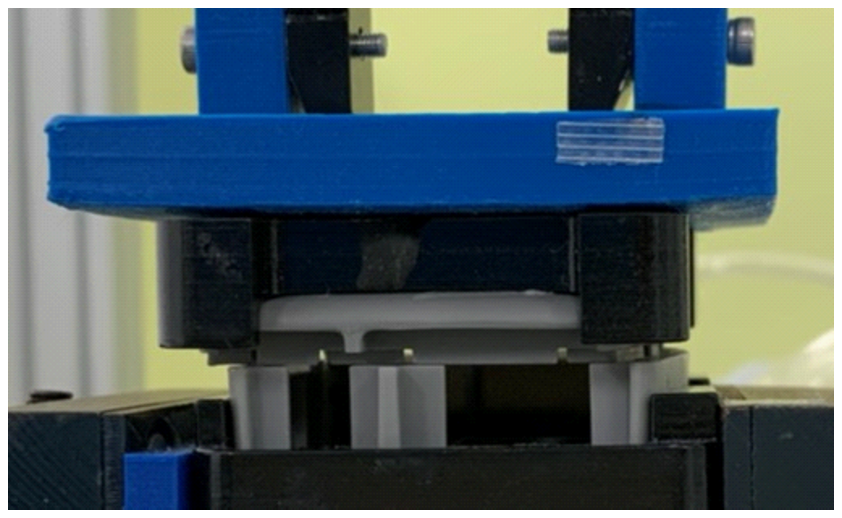}
  \subcaption{Assembly motion\newline}
 \end{minipage}
 \begin{minipage}[b]{0.4\linewidth}
  \centering
  \includegraphics[width=\linewidth]{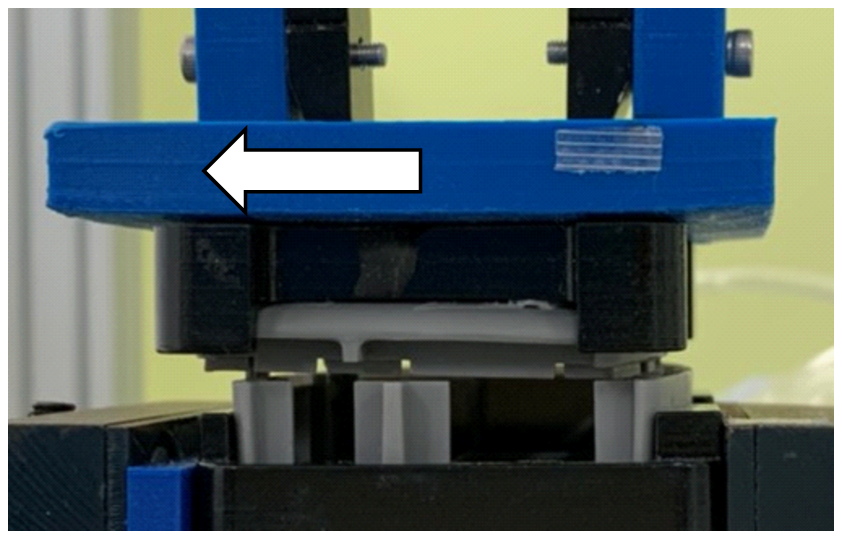}
  \subcaption{Additional probing in $+x$ direction}
 \end{minipage}\\
 \begin{minipage}[b]{0.4\linewidth}
  \centering
  \includegraphics[width=\linewidth]{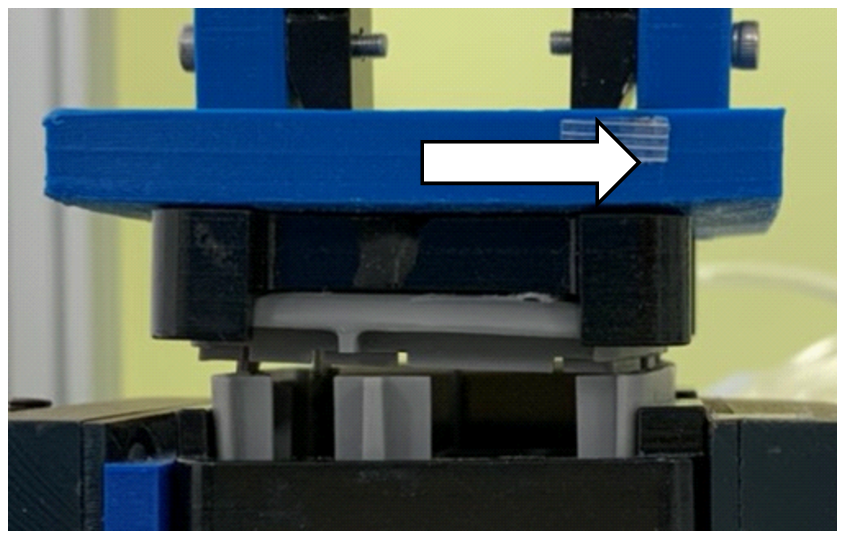}
  \subcaption{Additional probing in $-x$ direction}
 \end{minipage}
 \caption{Snapshots of snap assembly failure ($\Delta y=1.5[\mathrm{mm}]$, $\Delta \theta_z=1.5$[deg])}
   \label{snapshots failure}
   }
\end{figure}

Under the condition given in subsection \ref{4.1}, we collected both training and validation data. 
The obtained force/torque profiles of the characteristic components are shown in Figs. \ref{waveform Fz}, \ref{waveform Fx} and \ref{waveform Tx}. 
We can observe that the force/torque profile differ between the error states.  
Specifically, the force profile in the $z$ direction can be used to discriminate successful cases from failure cases. 
A successful case of assembly is shown in Fig. \ref{snapshots success}, where (a) and (b) show the male part's motion in the $+z$ direction and (c) and (d) show the additional probing in the $\pm x$ directions. 
The failure case of assembly is shown in Fig. \ref{snapshots failure}. 
% \ref{4.1}節の条件下で，スナップ工程，探り動作工程それぞれに対して学習用，及び検証用の波形データを取得した．
% 得られた各力・トルク成分波形を
% 図\ref{waveform Fz}，\ref{waveform Ty}，\ref{waveform Tx}に，
% また，スナップ工程，左探り動作工程，右探り動作工程それぞれにおける連続写真を
% 図\ref{snapshots success}，
% \ref{snapshots failure}に示す．
% ここで，力・トルク成分波形については，成否パターンの特徴が顕著に現れた例として，
% スナップ工程からは，成否パターン(1)，及び(9)を持つスナップアセンブリの実行時の$z$軸方向にかかる力の波形データと，
% 成否パターン(3)，及び(7)を持つスナップアセンブリの実行時の$y$軸まわりのトルクの波形データを，
% 探り動作工程からは，右方向探り動作において成否パターン(6)，及び(9)を持つスナップアセンブリの実行時の$x$軸まわりのトルクの波形データのみを示す．
% また，連続写真については，作業の成功例として，
% $\Delta y=0[\mathrm{mm}]$，$\Delta \theta_z=0.5[^\circ]$の場合と，
% 失敗例として，$\Delta y=1.5[\mathrm{mm}]$，$\Delta \theta_z=1.5[^\circ]$の場合のみ示し，図中の矢印は探り動作時のロボットハンドの移動方向を示す．

% \begin{figure}[!h]
%  \begin{minipage}[b]{0.49\linewidth}
%   \centering
%   \includegraphics[width=\linewidth]{picture/chap4/waveform/success_Fz_1stmove.eps}
%   \subcaption{Successful case (1)}
%  \end{minipage}
%  \begin{minipage}[b]{0.49\linewidth}
%   \centering
%   \includegraphics[width=\linewidth]{picture/chap4/waveform/failure6_Fz_1stmove.eps}
%   \subcaption{Error state (6)}
%  \end{minipage}
%  \caption{Force in $z$ direction corresponding to  successful/failure cases}
%     \label{waveform Fz}
% \end{figure}

% \begin{figure}[!h]
%  \begin{minipage}[b]{0.49\linewidth}
%   \centering
%   \includegraphics[width=\linewidth]{picture/chap4/waveform/failure7_Fx_1stmove.eps}
%   \subcaption{Error state (7)}
%  \end{minipage}
%  \begin{minipage}[b]{0.49\linewidth}
%   \centering
%   \includegraphics[width=\linewidth]{picture/chap4/waveform/failure8_Fx_1stmove.eps}
%   \subcaption{Error state (8)}
%  \end{minipage}
%  \caption{Force in $x$ direction corresponding to different error states}
%     \label{waveform Fx}
% \end{figure}

% \begin{figure}[!h]
%  \begin{minipage}[b]{0.49\linewidth}
%   \centering
%   \includegraphics[width=\linewidth]{picture/chap4/waveform/failure2_Tx_rgtsearch.eps}
%   \subcaption{Error state (2)}
%  \end{minipage}
%  \begin{minipage}[b]{0.49\linewidth}
%   \centering
%   \includegraphics[width=\linewidth]{picture/chap4/waveform/failure3_Tx_rgtsearch.eps}
%   \subcaption{Error state (3)}
%  \end{minipage}
%  \caption{Torque about $x$ axis corresponding to different error states during the additional probing in the $+x$ direction}
%     \label{waveform Tx}
% \end{figure}

%
% 図\ref{waveform Fz}，
% \ref{waveform Ty}，
% \ref{waveform Tx}の波形データについて，
% 波形の終端値や反転性において成否パターンそれぞれの特徴が確認できる．
% また他の成否パターンにおける各力・トルク成分波形からも同様の傾向が確認された．
% よって，以降の学習に有用な力情報が取得されていると考えられる．

\subsection{Feature extraction}
\label{4.3}

\begin{figure}[t]
  \begin{center}
    \begin{tabular}{c}
      \begin{minipage}{0.5\hsize}
        \begin{center}
          \includegraphics[width=\linewidth]{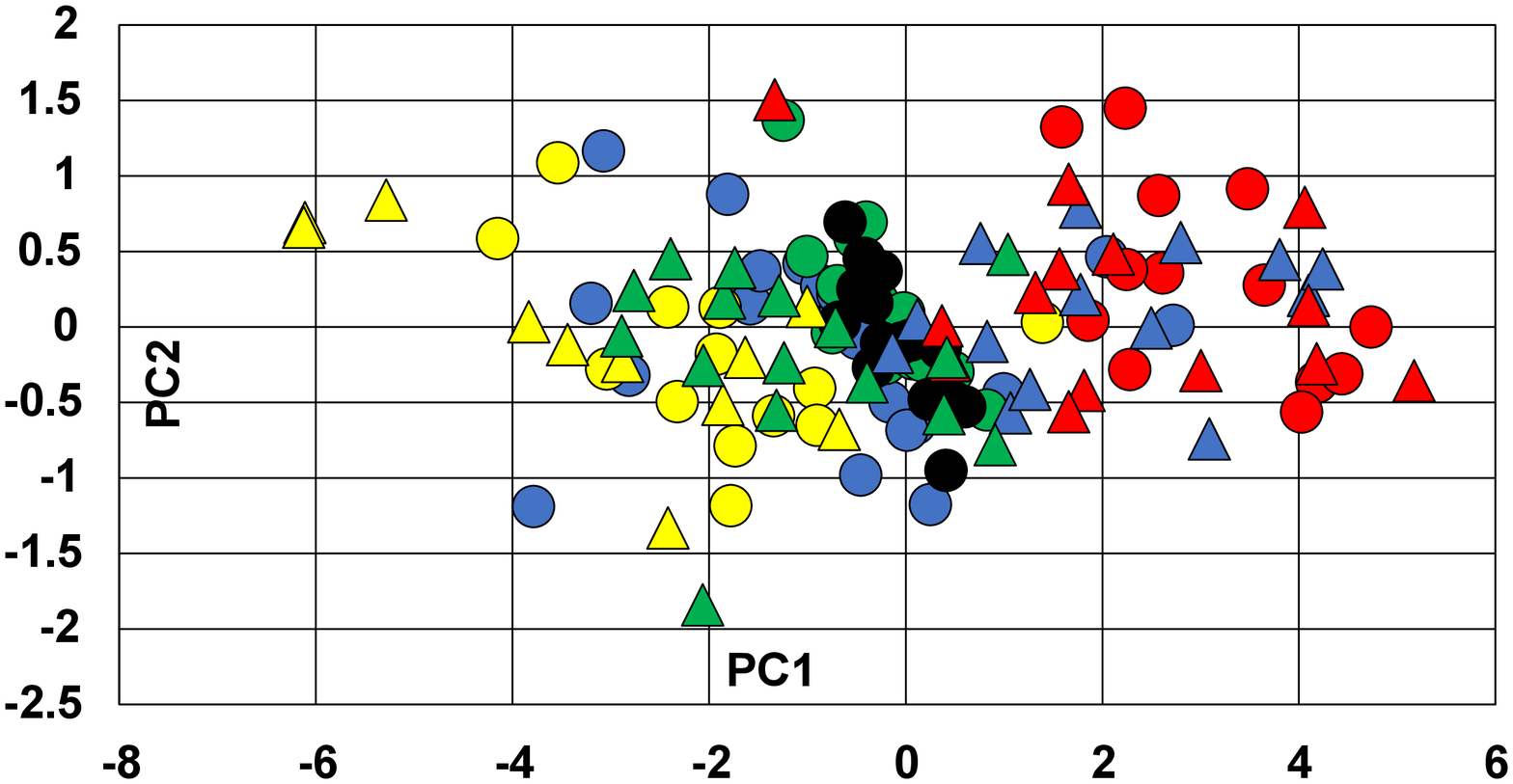}
          \hspace{1.6cm} (a)Force in $x$ direction
        \end{center}
      \end{minipage}
      \begin{minipage}{0.5\hsize}
        \begin{center}
          \includegraphics[width=\linewidth]{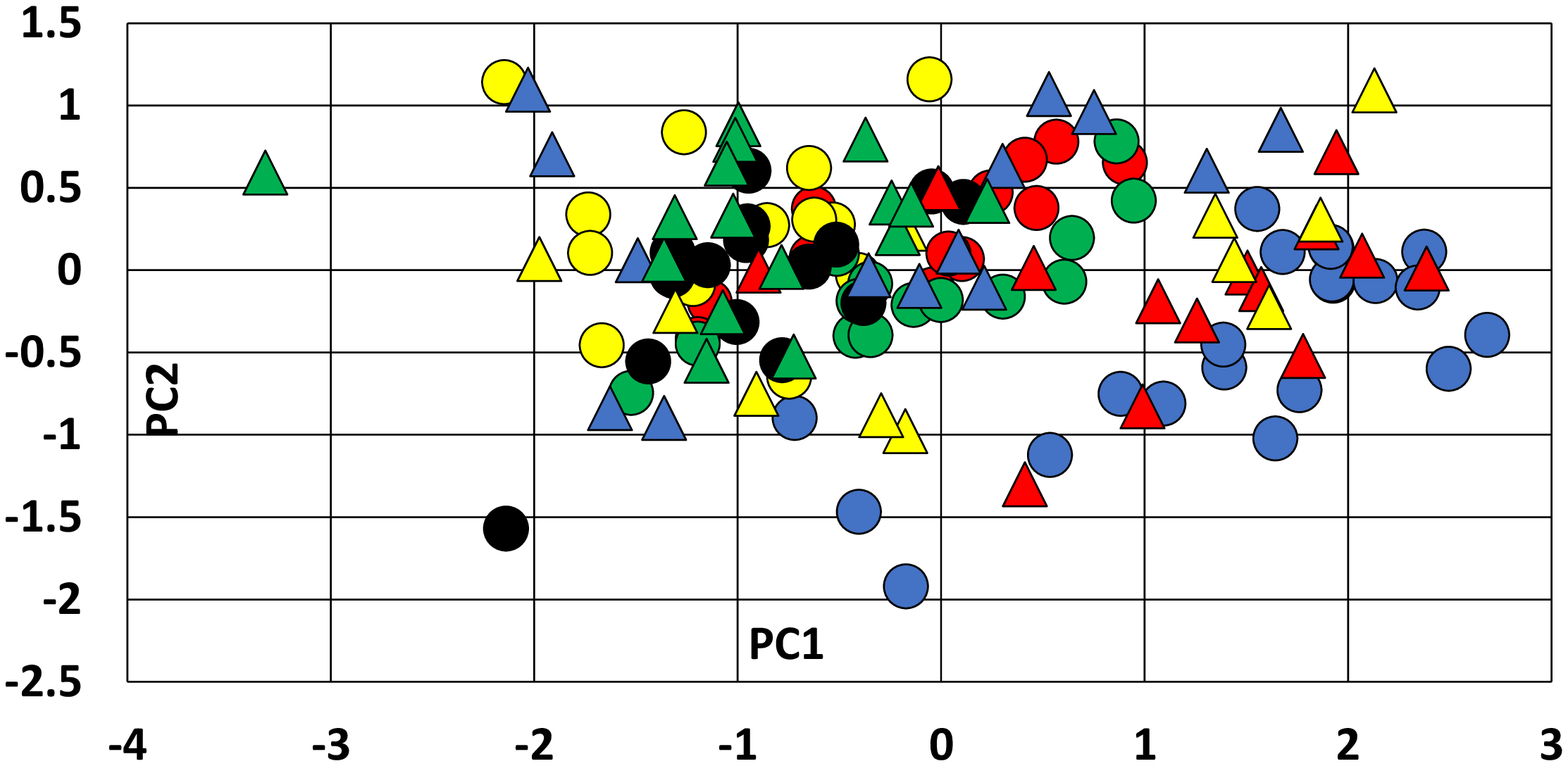}
          \hspace{1.6cm} (b)Force in $y$ direction
        \end{center}
      \end{minipage}\\\\
     \begin{minipage}{0.5\hsize}
        \begin{center}
          \includegraphics[width=\linewidth]{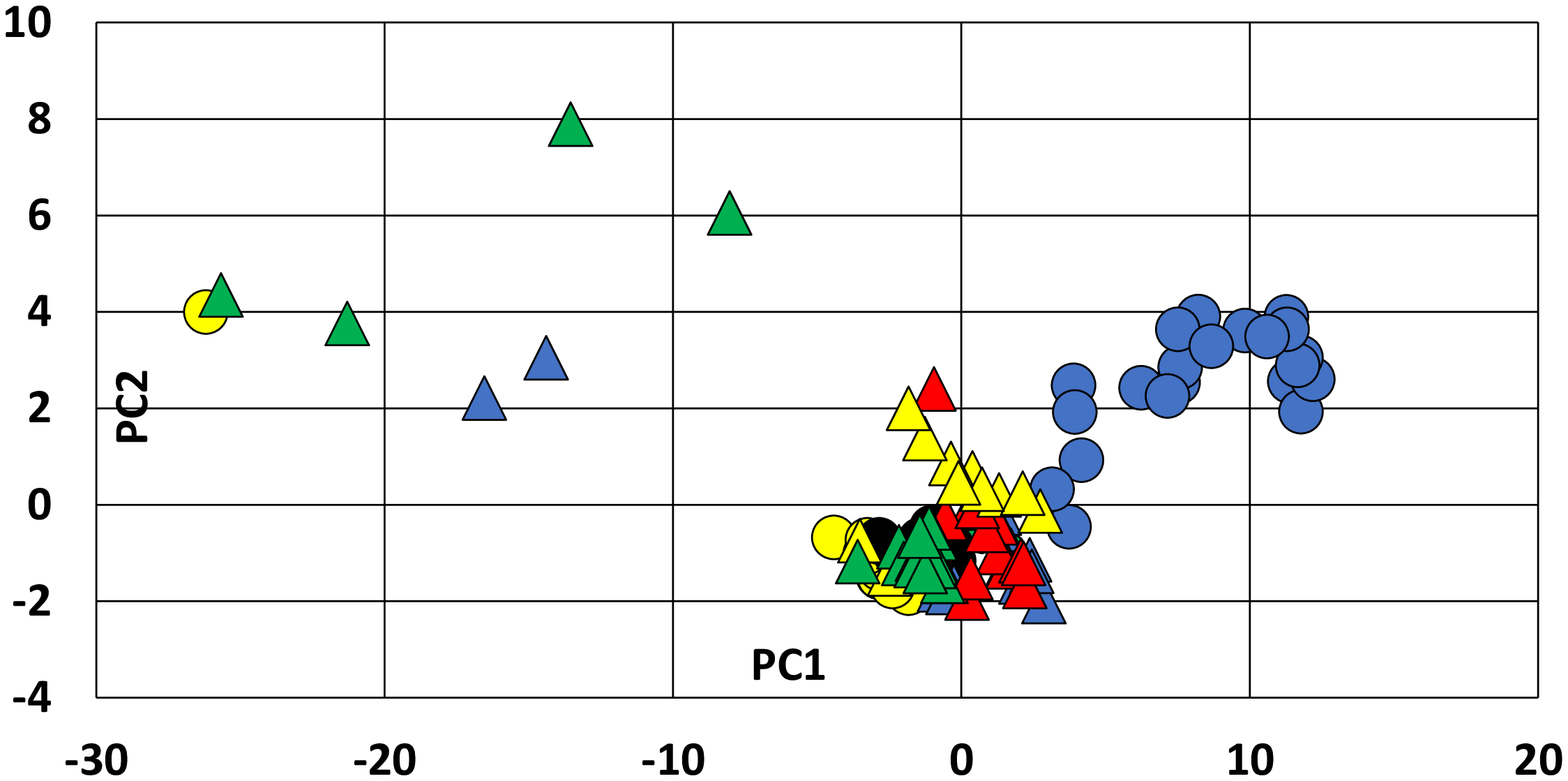}
          \hspace{1.6cm} (c)Force in $z$ direction
        \end{center}
      \end{minipage}
      \begin{minipage}{0.5\hsize}
        \begin{center}
          \includegraphics[width=\linewidth]{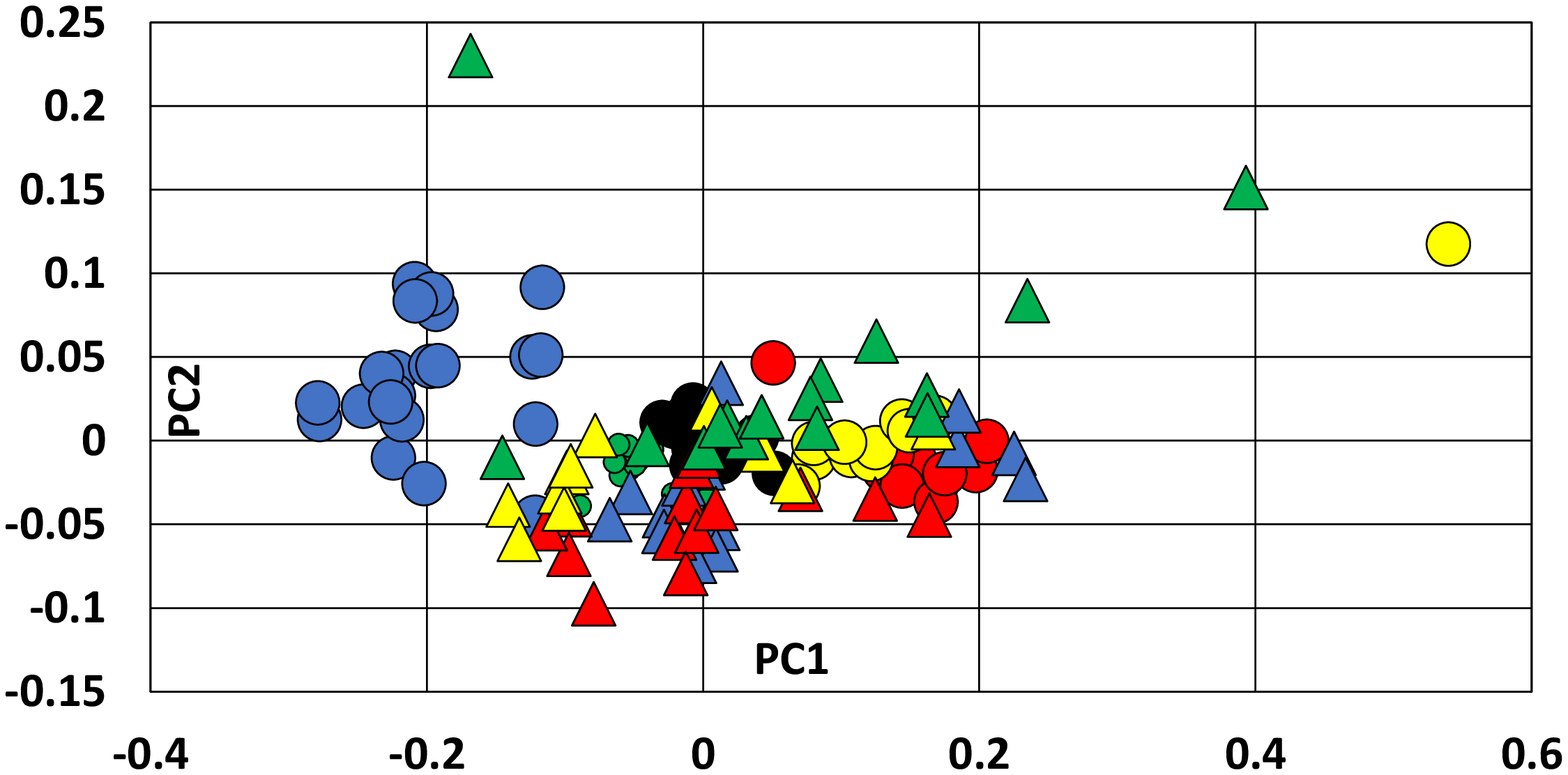}
          \hspace{1.6cm} (d)Torque about $x$ axis
        \end{center}
      \end{minipage}\\\\ %\clearpage \newpage
      \begin{minipage}{0.5\hsize}
        \begin{center}
          \includegraphics[width=\linewidth]{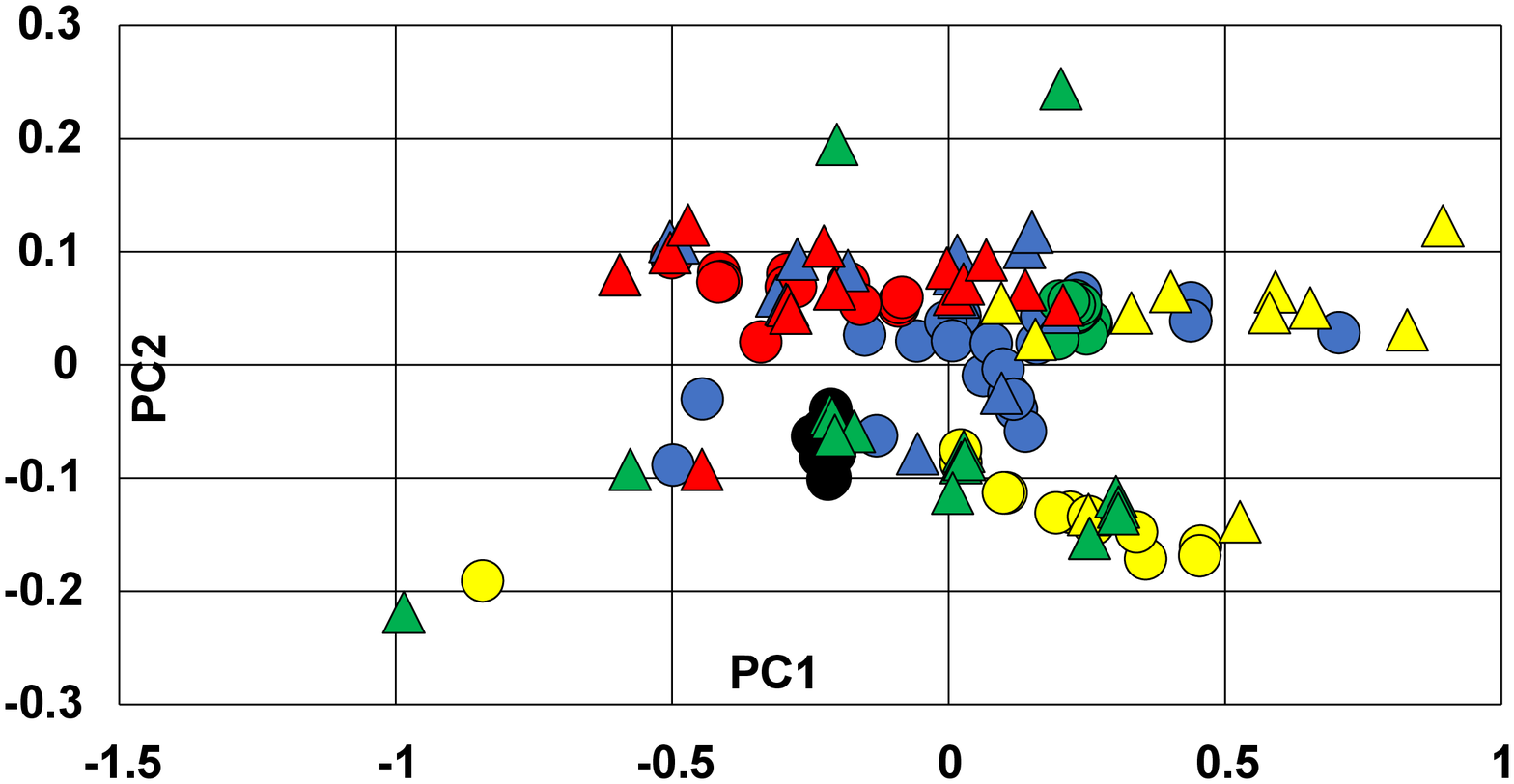}
          \hspace{1.6cm} (e)Torque about $y$ axis
        \end{center}
      \end{minipage}
      \begin{minipage}{0.5\hsize}
        \begin{center}
          \includegraphics[width=\linewidth]{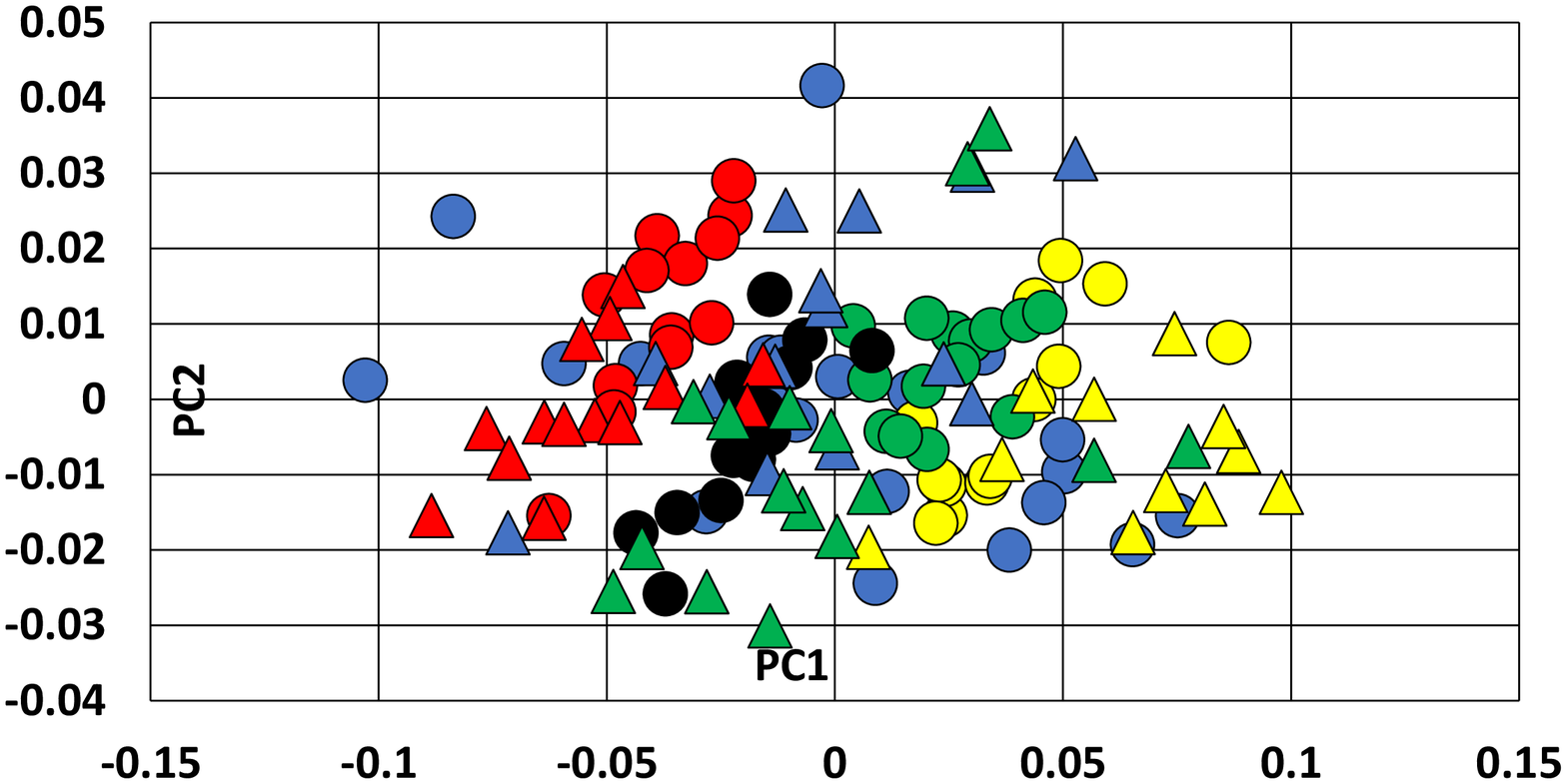}
          \hspace{1.6cm} (f)Torque about $z$ axis
        \end{center}
      \end{minipage}
    \end{tabular}
    \caption{Plots of functional principal component scores during snap assembly terminated at $t_{span}=2.0[\mathrm{s}]$, where blue, red, yellow, green and black circles, and blue, red, yellow and green triangles denote the success/error states (1), $\cdots$, (9), respectively.  }
    \label{feature quantity plot}
  \end{center}
\end{figure}

From a given force/torque profile of a snap assembly, we extracted the force/torque profile terminated at time $t_{span}$. Then, for the extracted force/torque profile terminated at $t_{span}$, we calculated the feature vector. 
%We checked if we can set $t_{span}$ before actually error occurs. 
Fig. \ref{feature quantity plot} shows the plot of the 1st and 2nd functional principal component scores assuming $t_{span}=2.0$[s] (before the error actually occurs). 
By observing this plot, some error states seem to be classified depending on the force/torque component, e.g., the blue circle of the force in the $z$-dierction. 
To form the decision tree, we will try to find such force/torque component that can accurately classify the error state. 
In our experiment, with the 1st and 2nd functional principal components, the contribution rate exceeded 90\%. 

% \ref{4.2}節で得られた波形データに対して，作業開始時刻，及び関数主成分分析に用いるデータ区間の終了時刻をそれぞれ$t_{start}=0[\mathrm{s}]$，$t_{span}[\mathrm{s}]$として$t_{span}$を様々に変更しながら\ref{3.3_sec:feature_extraction}節で述べた手順に従って関数主成分分析を行い，関数主成分得点を特徴量として抽出した．
% 本実験では，スナップ工程では$t_{span}$を$t_{span}=1.8[\mathrm{s}]$，$t_{span}=1.9[\mathrm{s}]$，及び，スナップ工程の終了時である$t_{span}=2.0[\mathrm{s}]$において，
% %，右探り動作工程，左探り動作工程では$t_{span}$を$t_{span}=[s]$，$t_{span}=[s]$，及び各探り動作工程の終了時である$t_{span}=[s]$において
% 右探り動作工程，左探り動作工程では$t_{span}$を両探り動作工程の終了時である
% $t_{span}=1.1[\mathrm{s}]$において
% 特徴量を抽出した．
% 以下に，スナップ工程で取得されたデータに対して$t_{span}=2[\mathrm{s}]$において抽出された特徴量分布図\ref{feature quantity plot}を示す．
% ここで，特徴量分布図\ref{feature quantity plot}
% について，本実験データでは，関数主成分分析の結果，6つの力・トルク成分すべてにおいて，第二主成分関数までで十分な寄与率(90\%以上)が達成されていたことや，特徴量をプロットした際の見やすさを考慮して，関数主成分得点についても第二関数主成分得点までを算出している．
% また，各グラフにおいて，
% 青色〇は成否パターン(1)を，
% 赤色〇は成否パターン(2)を，
% 黄色〇は成否パターン(3)を，
% 緑色〇は成否パターン(4)を，
% 黒色〇は成否パターン(5)を，
% 青色\triangle{}は成否パターン(6)を,
% 赤色\triangle{}は成否パターン(7)を,
% 黄色\triangle{}は成否パターン(8)を,
% 緑色\triangle{}は成否パターン(9)と表す．
%\clearpage

\color{black}
   
\subsection{Construction of the decision tree}
\label{4.4}

\begin{figure}[t]
  \begin{center}
          \includegraphics[width=\linewidth]{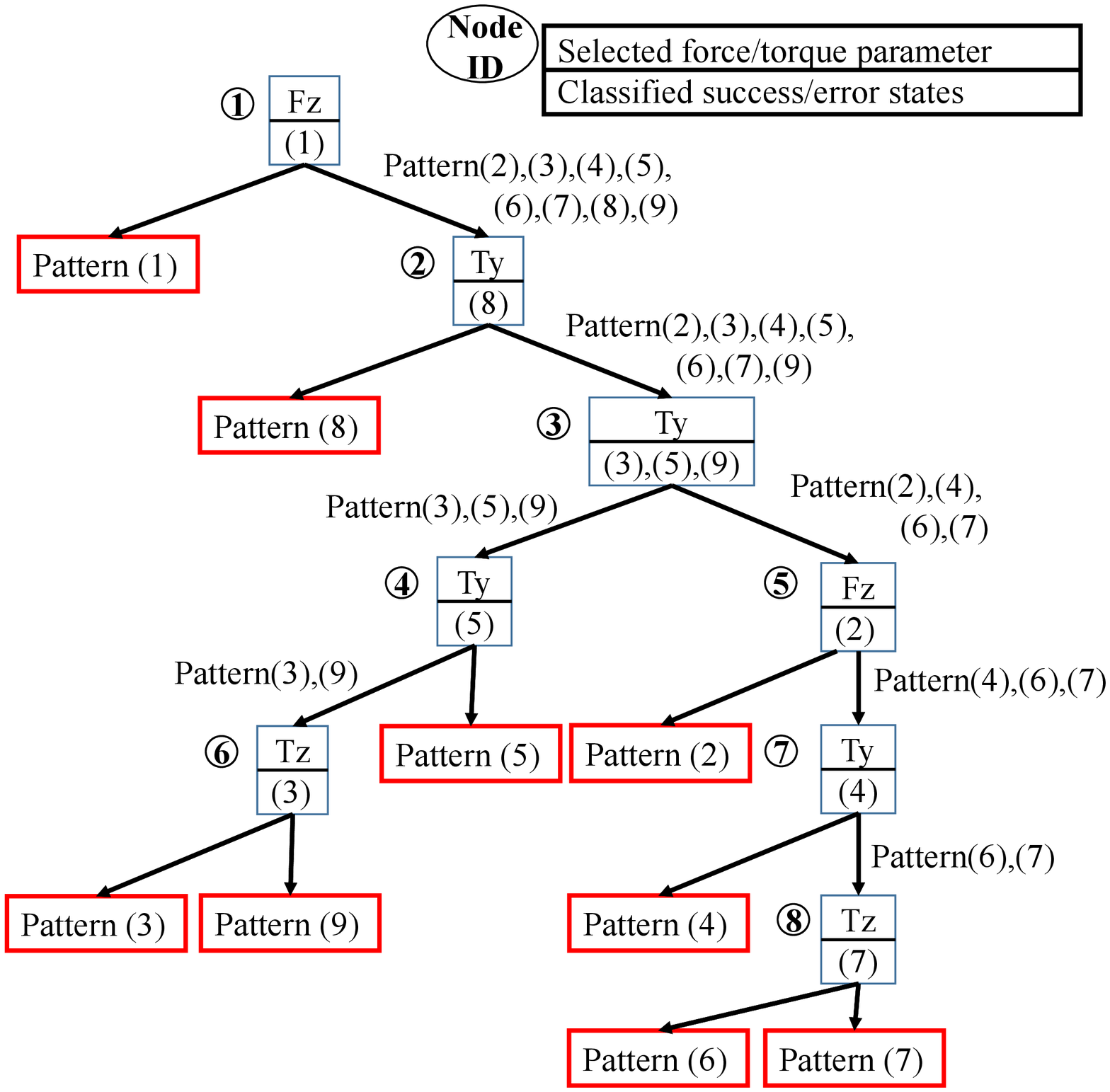}
          \caption{Decision tree of the snap assembly terminated at  $t_{span}=2.0[\mathrm{s}]$}
          \label{decision tree}
        \end{center}
\end{figure}

\begin{figure*}[t]
{\centering
 \begin{minipage}[b]{0.3\linewidth}
  \centering
  \includegraphics[width=\linewidth]{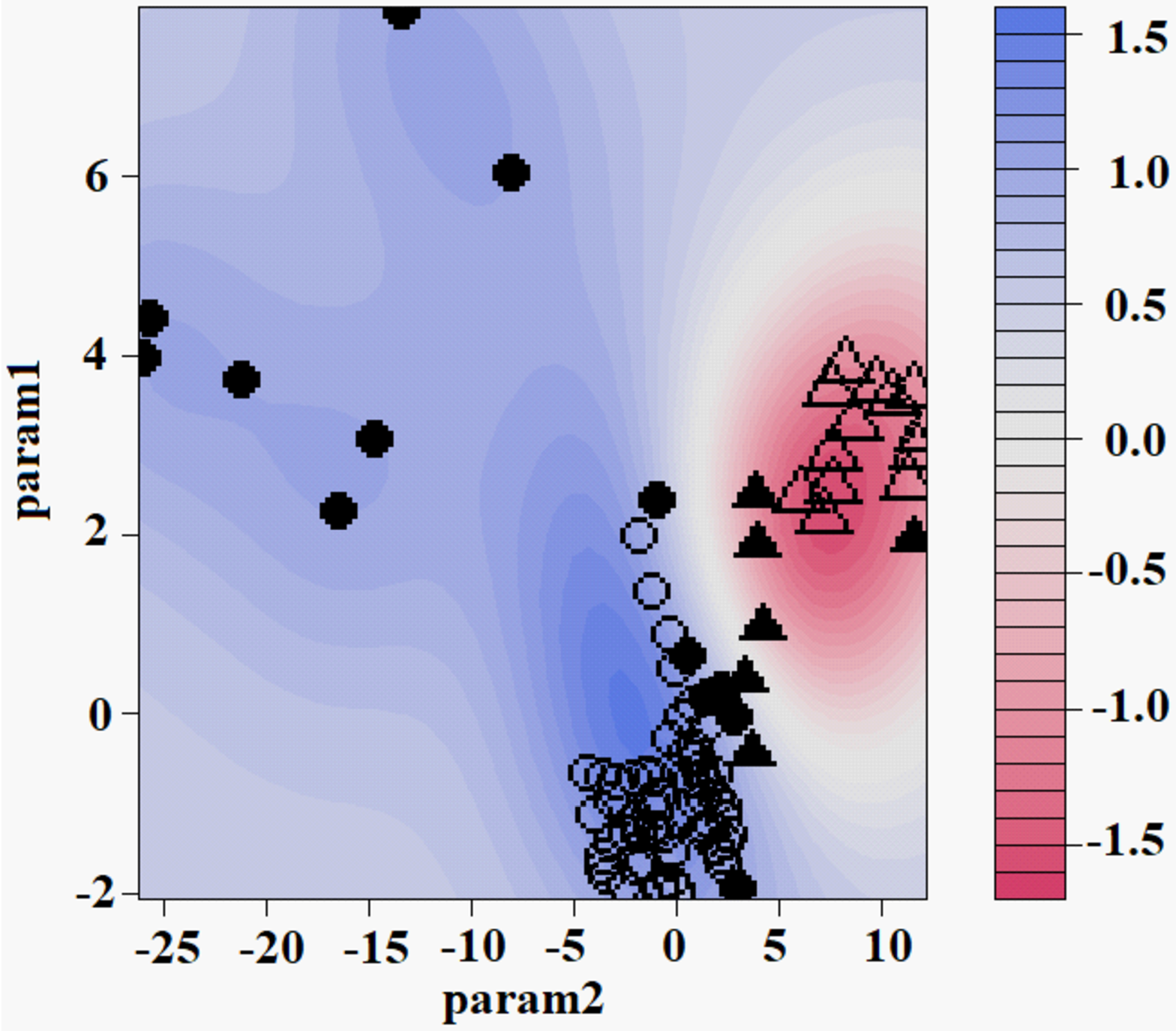}
  \subcaption{Node 1(Force in $z$ direction)}
 \end{minipage}
 \begin{minipage}[b]{0.3\linewidth}
  \centering
  \includegraphics[width=\linewidth]{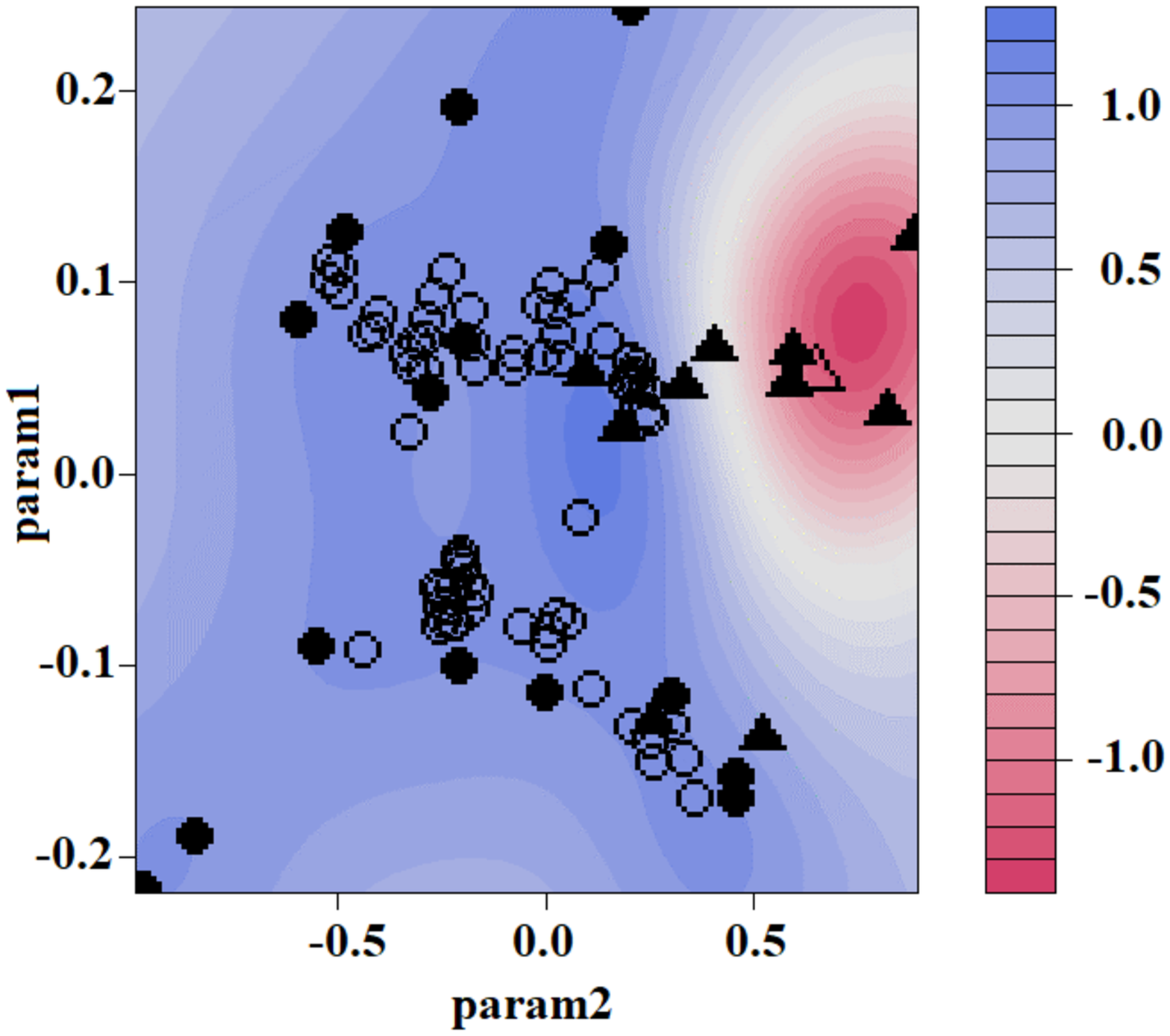}
  \subcaption{Node 2(Torque about $y$ axis)}
 \end{minipage}%\\
%\end{figure}
%\begin{figure}[H]
 \begin{minipage}[b]{0.3\linewidth}
  \centering
  \includegraphics[width=\linewidth]{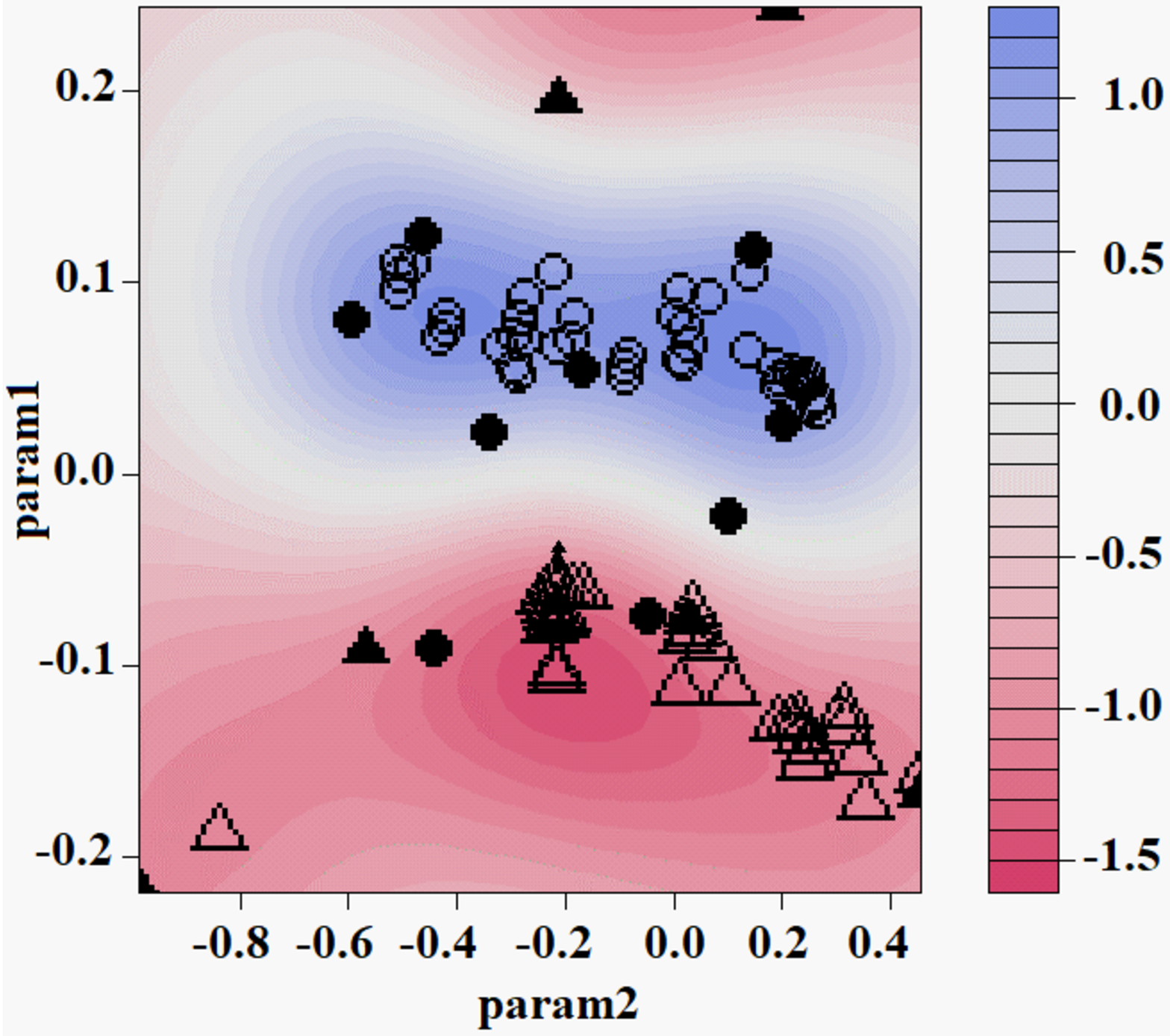}
  \subcaption{Node 3(Torque about $y$ axis)}
 \end{minipage}\\
 \begin{minipage}[b]{0.3\linewidth}
  \centering
  \includegraphics[width=\linewidth]{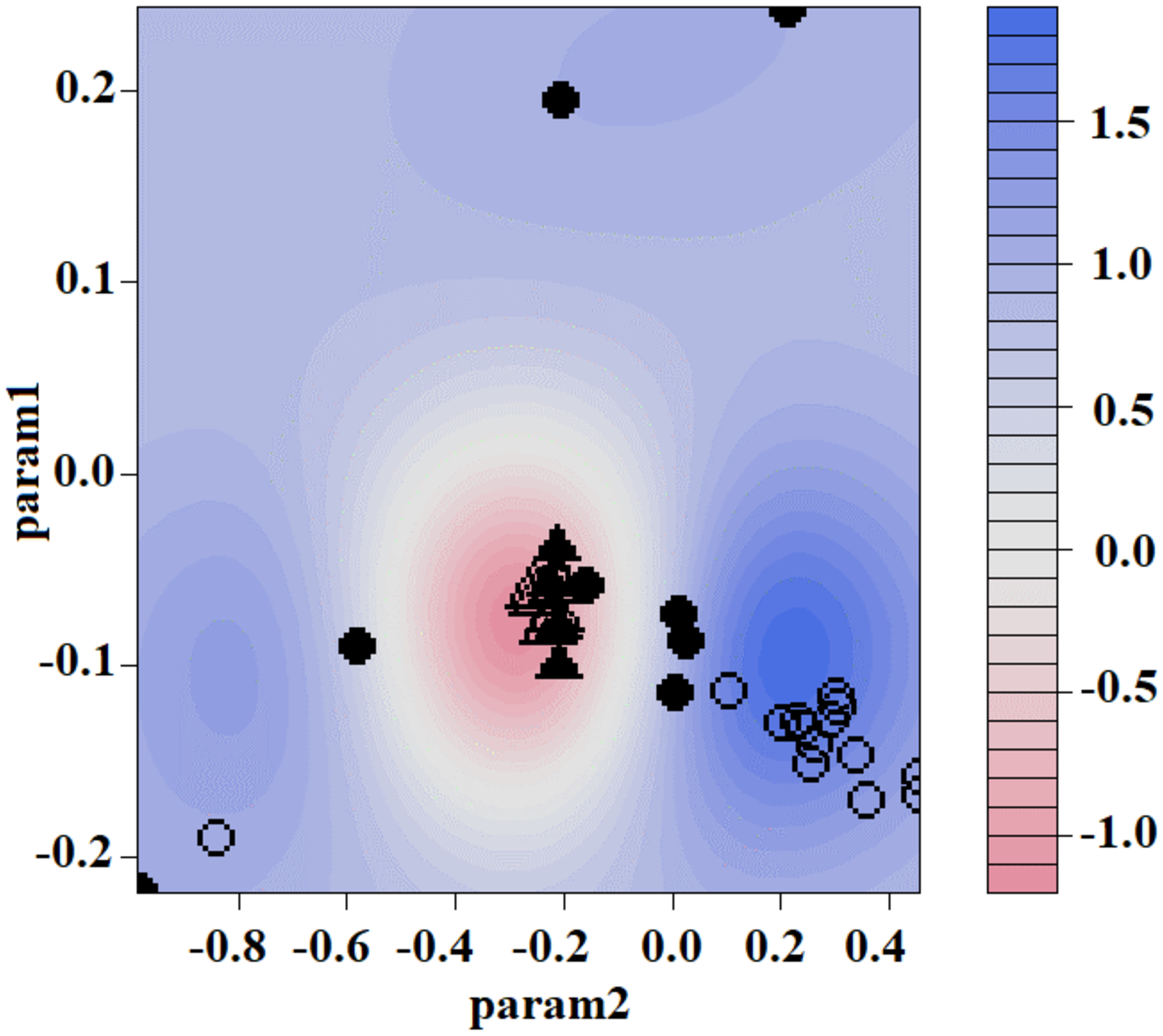}
  \subcaption{Node 4(Torque about $y$ axis)}
 \end{minipage}
% \end{figure}
% \setcounter{figure}{12}
%\begin{figure}[t]
 \begin{minipage}[b]{0.3\linewidth}
  \centering
  \includegraphics[width=\linewidth]{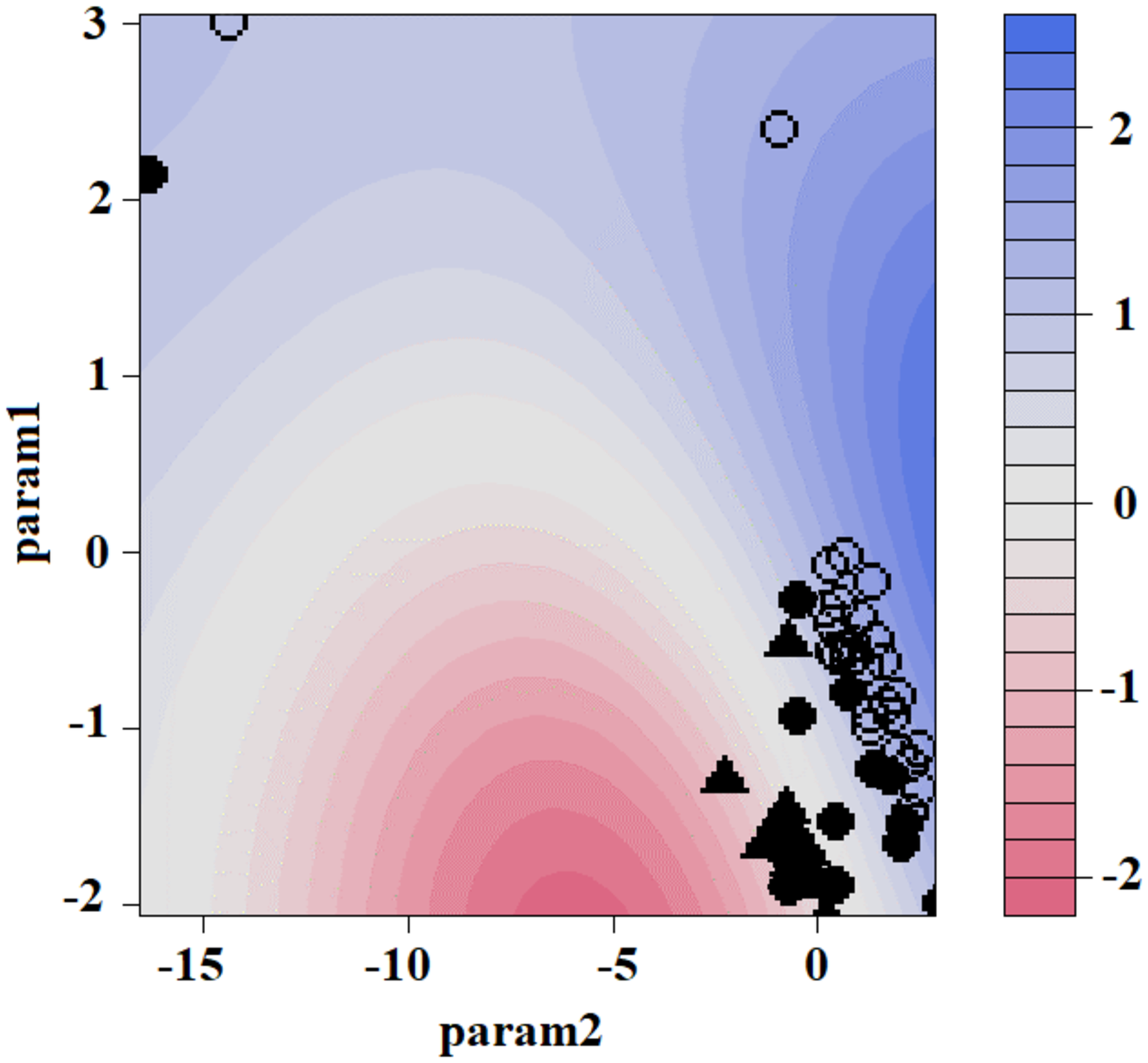}%\\
  \subcaption{Node 5(Force in $z$ direction)}
%(e) Node 5(Force in $z$ direction)
 \end{minipage}
 \begin{minipage}[b]{0.3\linewidth}
  \centering
  \includegraphics[width=\linewidth]{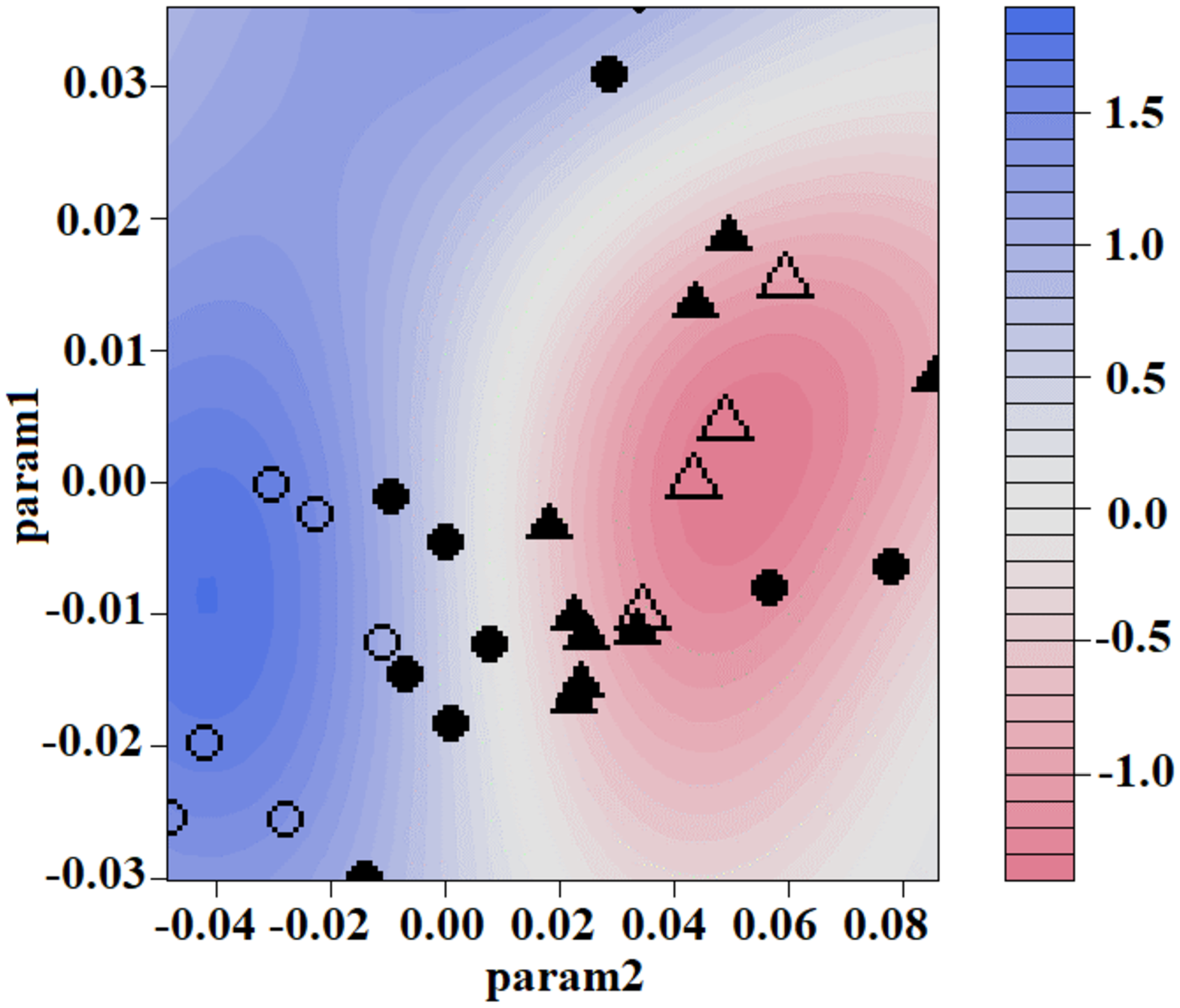}%\\
  \subcaption{Node 6(Torque about $z$ axis)}
%(f) Node 6(Torque about $z$ axis)
 \end{minipage}\\
%\end{figure}
%\begin{figure}[H]
 \begin{minipage}[b]{0.3\linewidth}
  \centering
  \includegraphics[width=\linewidth]{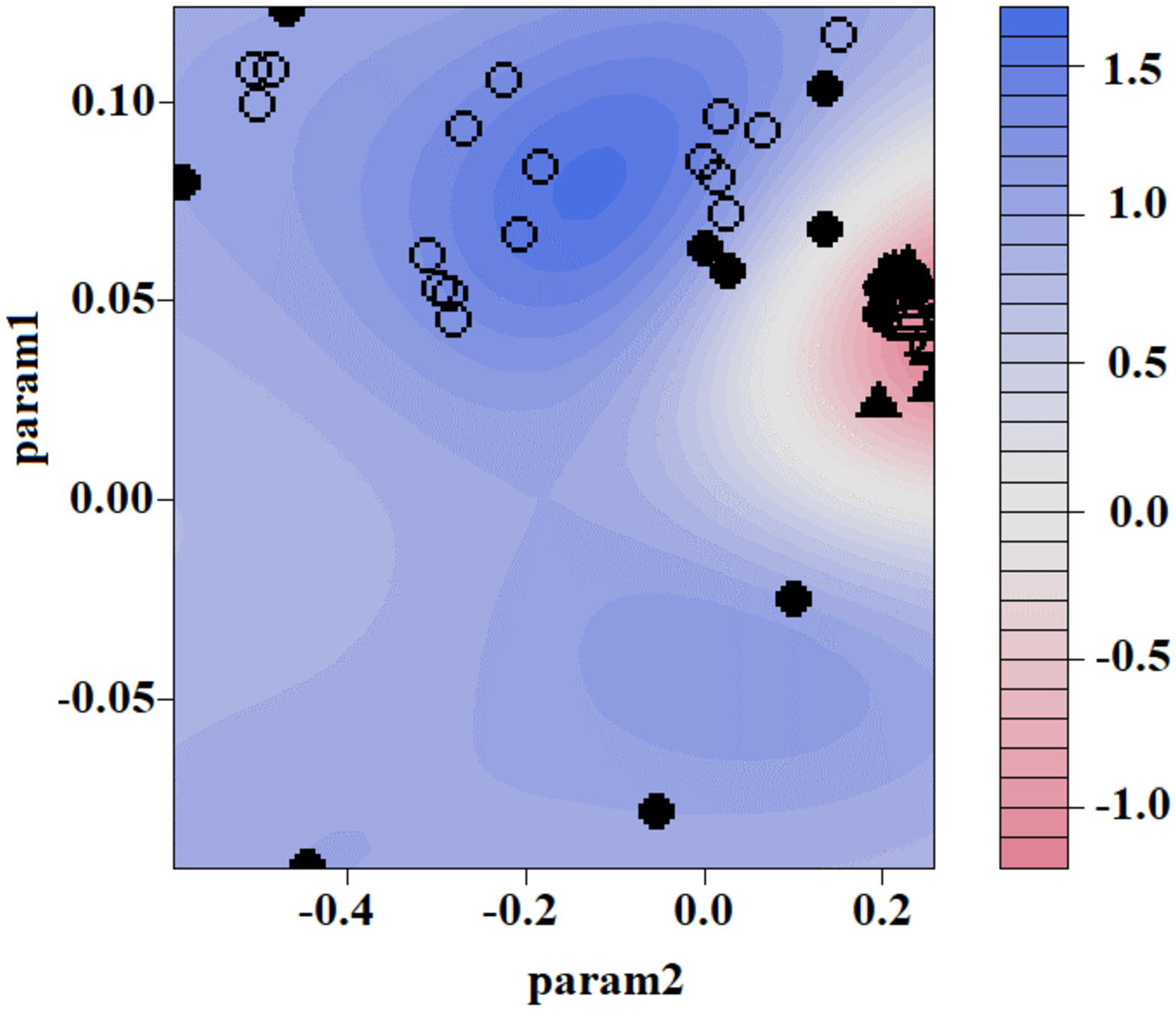}%\\
  \subcaption{Node 7(Torque about $y$ axis)}
%(g) Node 7(Torque about $y$ axis)
 \end{minipage}
 \begin{minipage}[b]{0.3\linewidth}
  \centering
  \includegraphics[width=\linewidth]{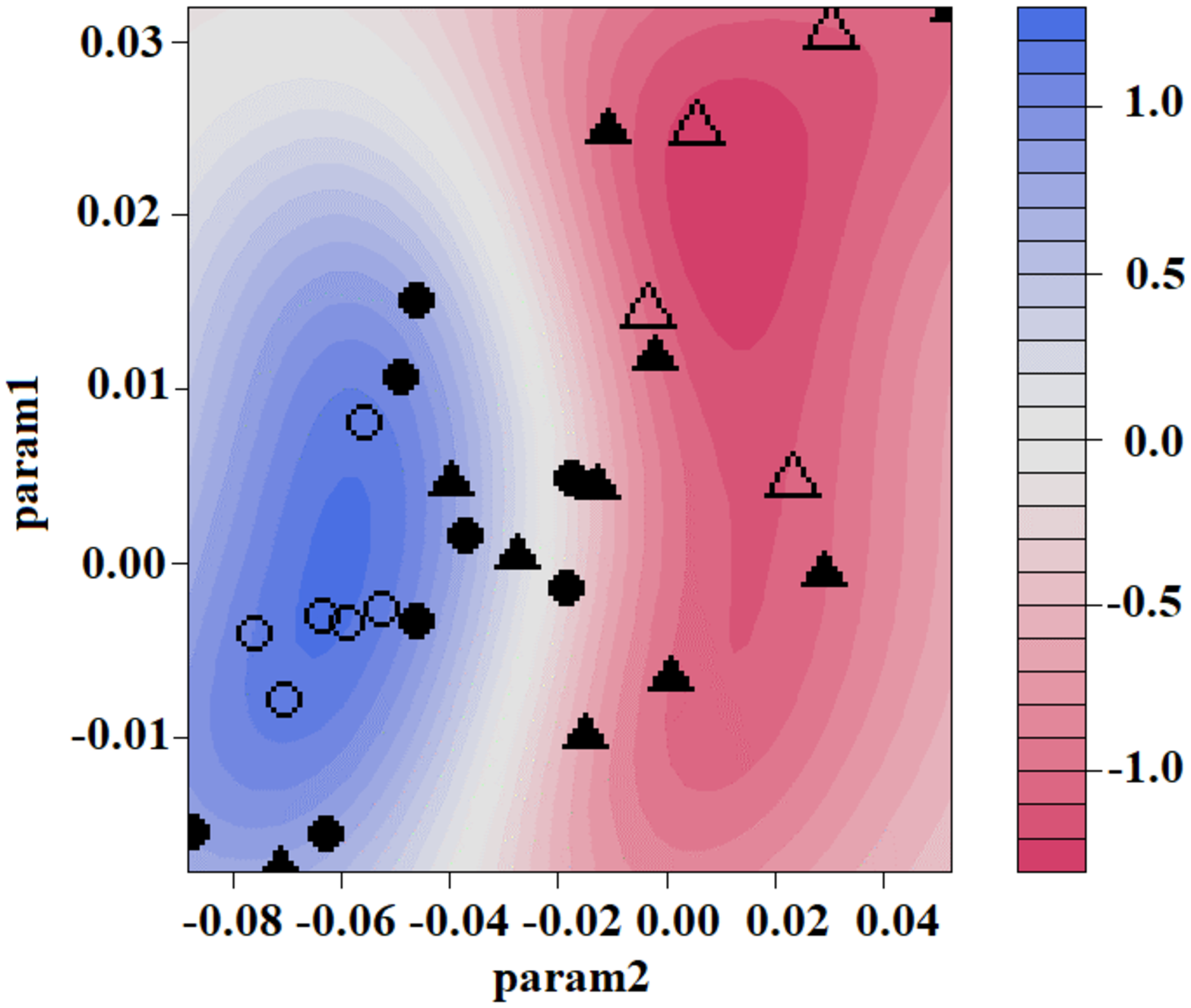}%\\
  \subcaption{Node 8(Torque about $z$ axis)}
%(h) Node 8(Torque about $z$ axis)
 \end{minipage}
 \caption{Classification result of error states by using SVM corresponding to feature quantities shown in Fig.\ref{decision tree}}
     \label{classifier}
     }
\end{figure*}

We constructed a decision tree using the feature vector 
obtained in subsection \ref{4.3} with the leave-one-out cross validation.  
In each node of the tree, we classified the error state by using the SVM with the kernel function. 
For the case of $t_{span}=2.0[\mathrm{s}]$, the decision tree constructed and the feature vectors classified using the SVM are shown in Figs. \ref{decision tree} and \ref{classifier}, respectively. 

\newpage

% \ref{4.3}節で抽出された特徴量に対して，アルゴリズム\ref{algorithm_SVM}を用いて決定木を構築した(図\ref{decision tree})．
% また本実験では，
% 一個抜き交差検証(leave-one-out cross-validation)に従い，
% SVMによる分類を行った．
% 以下に，
% スナップ工程において$t_{span}=2.0[\mathrm{s}]$において抽出された特徴量に対して構築された決定木，及び，各ノードにおけるSVMの実行結果を
% 図\ref{decision tree}，図\ref{classifier}に示す．
% また，$t_{span}=1.8[\mathrm{s}]$，$t_{span}=1.9[\mathrm{s}]$，$t_{span}=2.0[\mathrm{s}]$において構築された各ノードでの式\ref{accuracy}に基づく分類精度を
% 表\ref{node accuracy}に示す．
% ここで，
% 図\ref{decision tree}について，
% 各ノードにはSVMが実行された力・トルク成分，及び分類された成否パターン群を示している．
% また図\ref{classifier}(a)$\sim$(h)について，
% 〇と\triangle{}は分類された成否パターン群それぞれに属する特徴量であり，
% ●，▲
%はサポートベクトルである．

%\begin{figure}[h]
%  \begin{center}
%          \includegraphics[width=\linewidth]{picture/chap4/SVM/decision_tree_1stmove.eps}
%          \caption{Decision tree of snap assembly terminated at  $t_{span}=2.0[\mathrm{s}]$}
%          \label{decision tree}
%        \end{center}
%\end{figure}

\begin{center}
\begin{table}[h]
\begin{center}
\caption{Classification accuracy (eq.(\ref{accuracy})) of error states calculated at each node of decision tree [\%]}
  \begin{tabular}{|c|c|c|c|} \hline
    Node number & $t_{span}=1.8$[s] & 1.9 [s] & 2.0 [s] \\ \hline \hline
    Node1 & ~95.8~ & ~95.5~ & ~98.0~ \\ \hline
    Node2 & ~94.1~ & ~96.1~ & ~95.9~ \\ \hline
    Node3 & ~95.2~ & ~94.8~ & ~95.5~ \\ \hline
    Node4 & ~85.5~ & ~91.2~ & ~85.7~ \\ \hline
    Node5 & ~91.1~ & ~90.8~ & ~91.9~ \\  \hline
    Node6 & ~86.4~ & ~86.7~ & ~84.2~ \\ \hline
    Node7 & ~90.5~ & ~91.9~ & ~89.9~ \\ \hline
    Node8 & ~90.5~ & ~85.9~ & ~83.4~ \\ \hline
  \end{tabular}
  \label{node accuracy}
\end{center}
\end{table}
\end{center}

\noindent
The classification accuracies in the cases $t_{span}=1.8, 1.9$ and $2.0[\mathrm{s}]$ are shown in Table \ref{node accuracy}. 
The table indicates that 
the classification accuracy exceeds 85\% in most cases. 
%The accuracy is relatively low in the nodes 7 and 8 of 
%$t_{span}=1.8$ and $2.0[\mathrm{s}]$. 
Specifically, the accuracy is relatively high in all the nodes of 
$t_{span}=1.9[\mathrm{s}]$. 
These results imply that, by selecting an appropriate $t_{span}$, error states can be estimated with the classification accuracy between 85 and 95 \% even before errors actually occur. 

% 表\ref{node accuracy}より，
% $t_{span}=1.8[\mathrm{s}]$までの波形データを用いた場合，
% ノード7，8において分類精度が低いことが確認できる．
% 一方，$t_{span}=1.9[\mathrm{s}]$以降の波形データを追加した場合，
% 各ノードにおいて分類精度が高いことが確認できた．
% よって，十分な識別精度を得るために，
% スナップ工程では
% $t_{span}=1.9[\mathrm{s}]$程度までの力情報が重要であり，
% $t_{span}=1.8[\mathrm{s}]$までの早い段階での力情報のみでは作業途中での識別が困難であると考えられる．

\subsection{Identification of error states}
\label{4.5}

By using the validation data obtained in subsection \ref{4.2}, 
we checked how early we can identify error states 
assuming $t_{span}=1.8, 1.9$ and $2.0 [\mathrm{s}]$. 
The validation data include 45 successful and 
40 error cases. The result of error-state identification for assembly motion when the male part moves in the $+z$ direction is presented in Table \ref{identify snap process}. 
The success rate of classification exceeds 87\%. 

Then, the result of classification when performing additional probing in the $\pm x$ direction is presented in Table \ref{identify search process}. 
The success rate of classification by additional probing itself is not very high. 
Furthermore, Table \ref{identify both processes} shows the success rate of classification achieved by additional probing in the $\pm x$ direction performed only when the classification accuracy of assembly motion is less than the threshold. 
In this case, the success rate of classification exceeds 92\%, which demonstrates the effectiveness of using additional probing. 
The result of classification of each error state corresponding to  $t_{span}=2.0[\mathrm{s}]$
is presented in Table \ref{identify detail}. 

% \ref{4.2}節で取得した検証用の波形データに対して，
% $t_{span}$を$t_{span}=1.8[\mathrm{s}]$，$t_{span}=1.9[\mathrm{s}]$，$t_{span}=2.0[\mathrm{s}]$として，それぞれのデータ区間に対して特徴量を抽出し，\ref{4.4}節で構築された対応するデータ区間における決定木を用いて成否パターンの識別を行うことで，どの程度早い段階で正しい識別が可能であるのかを検証した．
% 検証用のデータとして，
% 表\ref{identify detail}
% に示すように，成否パターン(1)については45回スナップアセンブリを，
% 成否パターン(2)$\sim$(9)については各5回スナップアセンブリを行い，
% 計85回分のスナップ工程，探り動作工程における波形データを取得し，成否パターンの識別を行った．
% 探り動作工程を導入せずにスナップ工程のみで識別を行った結果を
% 表\ref{identify snap process}に，
% スナップ工程での識別結果を用いずに探り動作工程のみで識別を行った結果を
% 表\ref{identify search process}に，
% \ref{3.5}節に従い，探り動作工程を導入した際の識別結果を
% 表\ref{identify both processes}に，
% 成否パターンそれぞれに対する識別結果を
% 表\ref{identify detail}
% に示す．ここで，表\ref{identify detail}では，$t_{span}=2.0[\mathrm{s}]$での識別結果のみを示す．
\begin{center}
\begin{table}[t]
\begin{center}
\caption{Classification result of success/error states in the assembly motion where the male part moves in the $+z$ direction}
  \begin{tabular}{|c|c|c|} \hline
  $t_{span}[\mathrm{s}]$ & Number of & Success rate(\%) \\
    & successful identification &  \\ \hline \hline
    1.8 & 74 & 87.1 \\ \hline
    1.9 & 84 & 98.8 \\ \hline
    2.0 & 83 & 97.6 \\ \hline
  \end{tabular}
  \label{identify snap process}
\end{center}
\end{table}
\end{center}

\begin{center}
\begin{table}[t]
\begin{center}
\caption{Classification result of success/error states with using the additional probing}
  \begin{tabular}{|c|c|c|} \hline
  Number of & Success rate(\%) \\
  successful identification &  \\ \hline \hline
    63 & 74.1 \\ \hline
  \end{tabular}
  \label{identify search process}
\end{center}
\end{table}
\end{center}

\begin{center}
\begin{table}[t]
\begin{center}
\caption{Classification result of success/error states with using the additional probing after the assembly motion where the male part moves in the $+z$ direction}
  \begin{tabular}{|c|c|c|c|c|} \hline
  $t_{span}[\mathrm{s}]$ & Improved & Deteriorated & Identification & Success\\
   & identification & identification & Succeeded & rate(\%) \\ \hline \hline
  1.8 & 9 & 4 & 79 & 92.9 \\ \cline{1-5}
  1.9 & 0 & 0 & 84 & 98.8 \\ \cline{1-5}
  2.0 & 2 & 0 & 85 & 100 \\ \hline
  \end{tabular}
  \label{identify both processes}
\end{center}
\end{table}
\end{center}

\begin{center}
\begin{table}[t]
\begin{center}
\caption{Result of identification for successful/error states calculated at $t_{span}=2.0[\mathrm{s}]$}
%\hspace*{-0.6cm}
  \begin{tabular}{|c|c|c|c|c|c|} \hline
  Success & $\Delta x$ & $\Delta \theta_z$ & Success rate & Success rate & Success rate\\
  /error & [$\mathrm{mm}$] & [deg] & (assembly) & (probing) & (probing after \\
   states  &  &  & [\%] & [\%] & assembly)[\%]\\ \hline \hline
    (1) & -0.25 & -0.25 & 80 & 80 & 100 \\ \cline{1-6}
    (1) & -0.5 & 0.5 & 100 & 100 & 100 \\ \cline{1-6}
    (1) & -0.5 & 0 & 100 & 100 & 100 \\ \cline{1-6}
    (1) & 0.25 & -0.25 & 100 & 20 & 100\\ \cline{1-6}
    (1) & 0.5 & 0.5 & 100 & 100 & 100 \\ \cline{1-6}
    (1) & 0.5 & 0 & 100 & 0 & 100 \\ \cline{1-6}
    (1) & 0 & -0.5 & 100 & 40 & 100 \\ \cline{1-6}
    (1) & 0 & 0.5 & 100 & 80 & 100 \\ \cline{1-6}
    (1) & 0 & 0 & 100 & 20 & 100 \\ \cline{1-6}
    (2) & 2 & 0 & 100 & 100 & 100 \\ \cline{1-6}
    (3) & -2 & 0 & 80 & 100 & 100 \\ \cline{1-6}
    (4) & 0 & 2 & 100 & 100 & 100 \\ \cline{1-6}
    (5) & 0 & -2 & 100 & 100 & 100 \\ \cline{1-6}
    (6) & 1.5 & 1.5 & 100 & 100 & 100 \\ \cline{1-6}
    (7) & 1.5 & -1.5 & 100 & 60 & 100 \\ \cline{1-6}
    (8) & -1.5 & 1.5 & 100 & 60 & 100 \\ \cline{1-6}
    (9) & -1.5 & -1.5 & 100 & 100 & 100 \\ \hline
  \end{tabular}
  \label{identify detail}
\end{center}
\end{table}
\end{center}

Table \ref{identify snap process} indicates that the accuracies of error-state identification 
were 87.1\%, 98.8\%, and 97.6\% when $t_{span}=1.8, 1.9$ and 2.0 [s], respectively. From this observation, without information included between $t_{span}=1.8$ and $1.9$[s] it becomes difficult to identify error states. $t_{span}$ should be set larger than 1.9 [s]. 
 
% 表\ref{identify snap process}より，スナップ工程での識別成功率は，
% データ区間$t_{span}[\mathrm{s}]=1.8$においては14.1\%，
% $t_{span}[\mathrm{s}]=1.9$においては62.4\%，
% $t_{span}[\mathrm{s}]=2.0$においては61.2$\%$であった．
% このことから，$t_{span}[\mathrm{s}]=1.8$\sim$1.9$の力情報が
% 作業部品間の物理作用を最も反映しており，成功パターンを識別するために重要であると考えられる．
% また，これら各区間は，スナップアセンブリが完了するよりも前の時点までの力情報であるため，提案手法による作業途中での成否パターンの識別可能性が示唆されたといえる．
% 特に，表\ref{identify detail}より，
% 成否パターン(1)の作業途中での識別精度は82.2\%と高く，スナップアセンブリの成功を迅速に検知可能である．
% 一方，表\ref{identify search process}，表\ref{identify both processes}より，
% 探り動作工程における識別精度はある程度高いことが確認されたにもかかわらず，
% 探り動作工程を導入した際の識別精度は改善されなかった．
% これについては，後述の考察で述べる．

\subsection{Error recovery}
\label{4.6}

\begin{figure}[t]
{\centering
 \begin{minipage}[b]{0.4\linewidth}
  \centering
  \includegraphics[width=\linewidth]{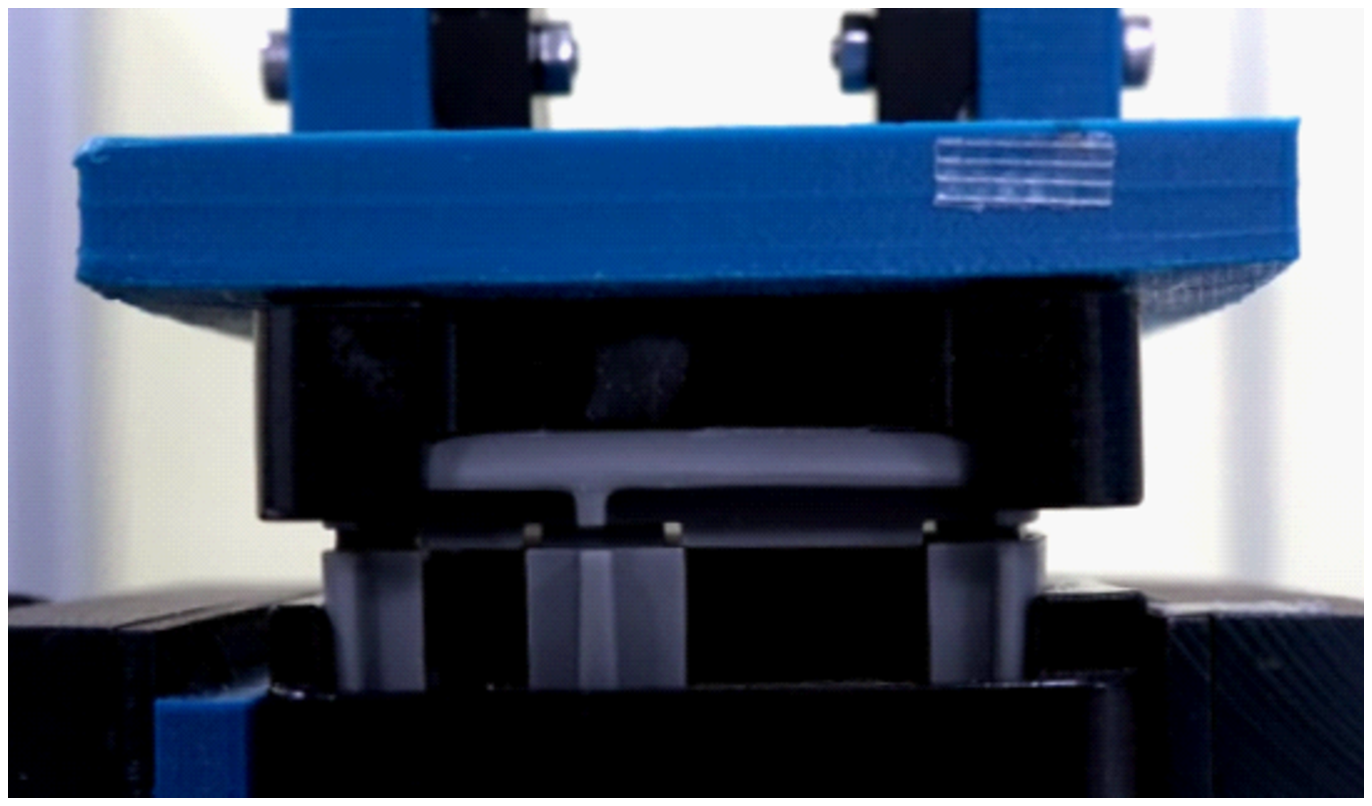}
  \subcaption{Snap insertion\newline}
 \end{minipage}
 \begin{minipage}[b]{0.4\linewidth}
  \centering
  \includegraphics[width=\linewidth]{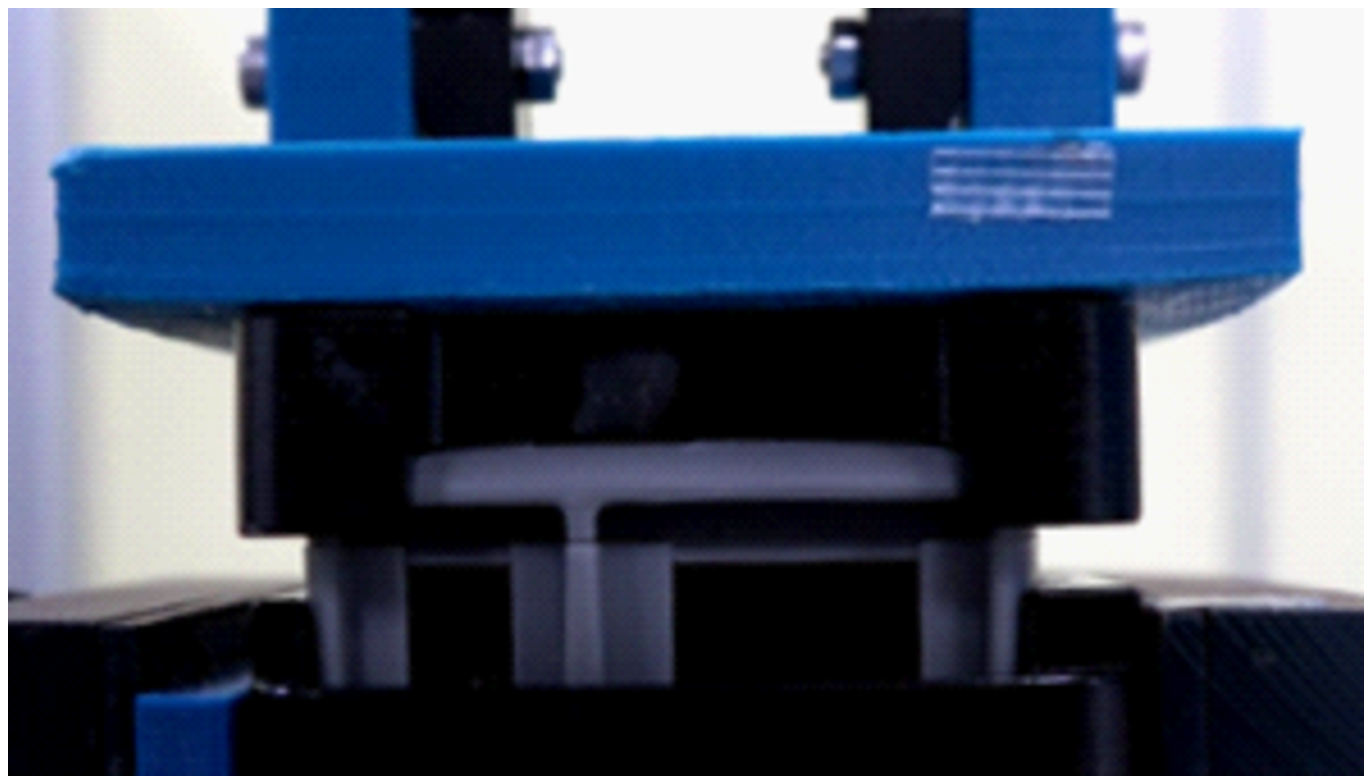}
  \subcaption{Snap assembly has successfully finished}
 \end{minipage}
 \caption{Error recovery from the offset pattern (1)}
    \label{recovery success}
    }
\end{figure}

\begin{figure}[t]
{\centering
 \begin{minipage}[b]{0.4\linewidth}
  \centering
  \includegraphics[width=\linewidth]{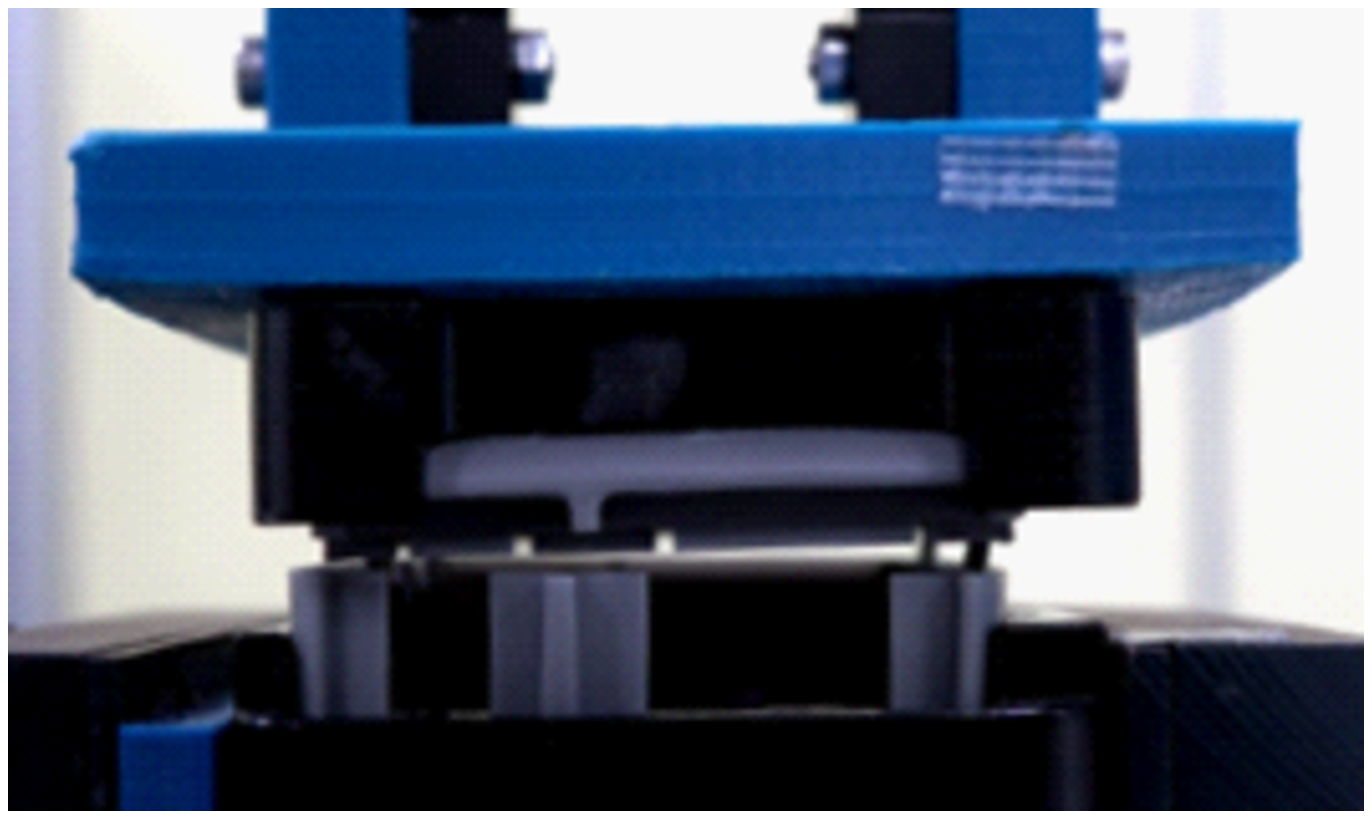}
  \subcaption{Snap insertion\newline}
 \end{minipage}
 \begin{minipage}[b]{0.4\linewidth}
  \centering
  \includegraphics[width=\linewidth]{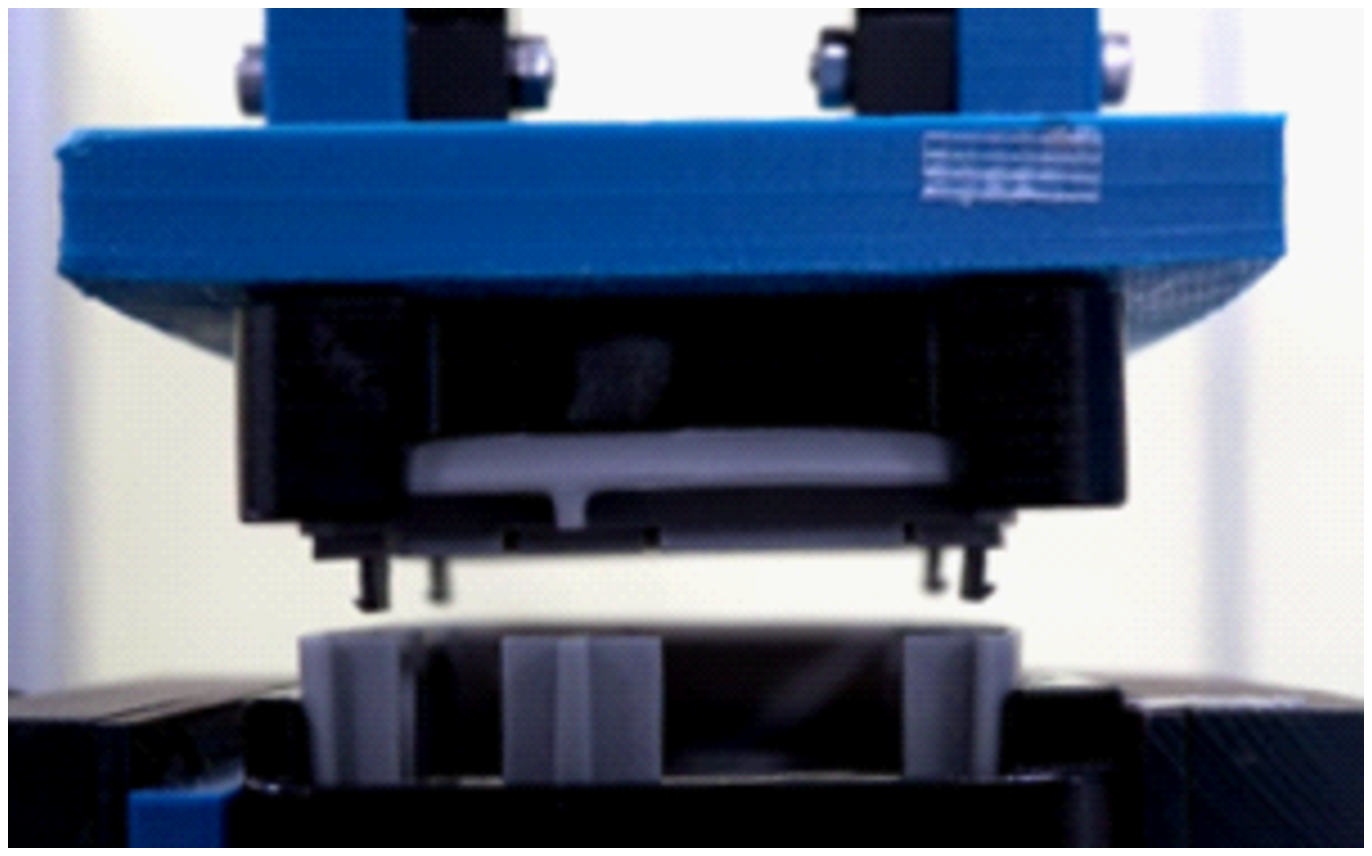}
  \subcaption{Error recovery motion\newline}
 \end{minipage}\\
 \begin{minipage}[b]{0.4\linewidth}
  \centering
  \includegraphics[width=\linewidth]{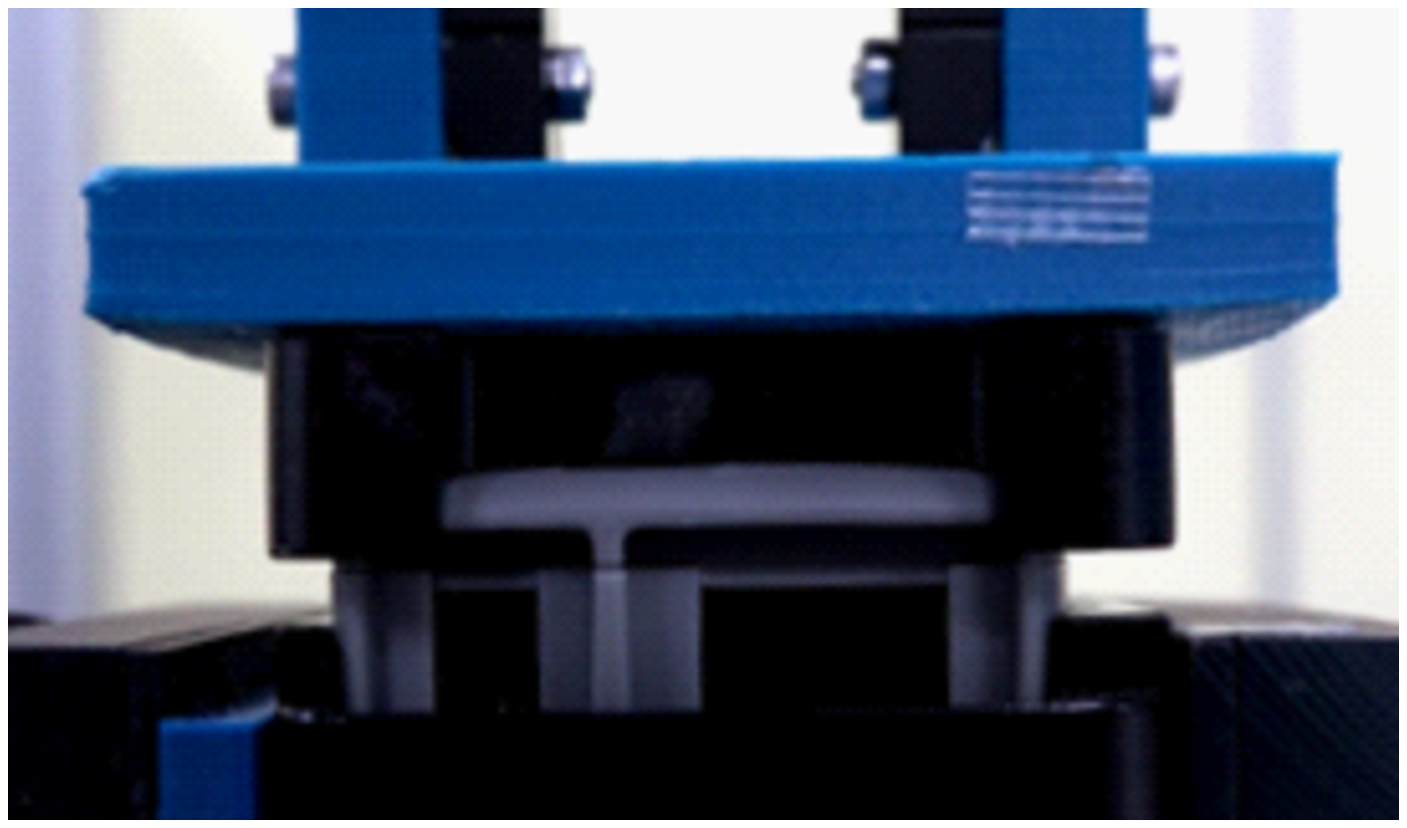}
  \subcaption{Snap assembly has successfully finished}
 \end{minipage}
 \caption{Error recovery from the offset pattern (2)}
    \label{recovery failure nosearch}
    }
\end{figure}

\begin{figure}[t]
{\centering
 \begin{minipage}[b]{0.4\linewidth}
  \centering
  \includegraphics[width=\linewidth]{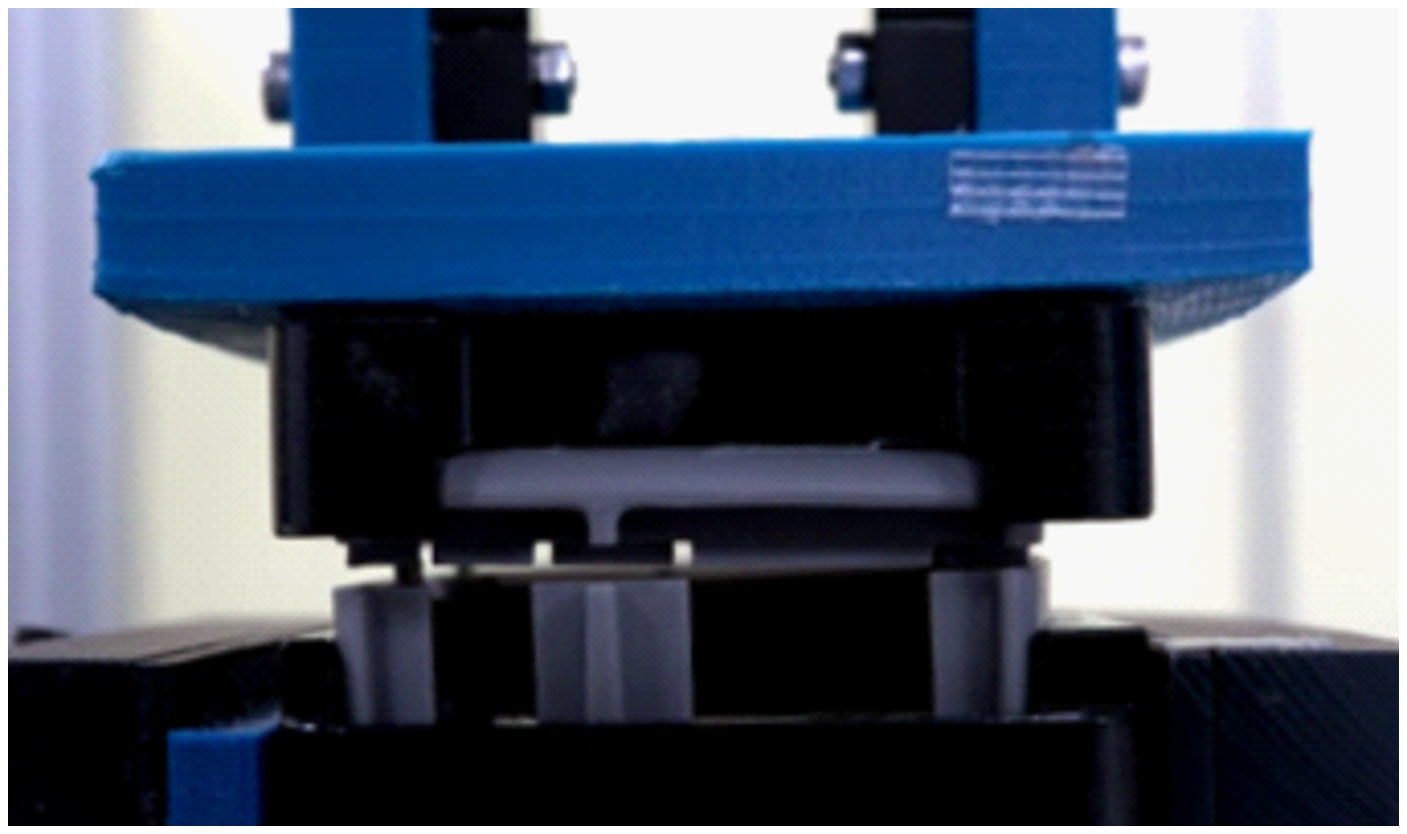}
  \subcaption{Assembly motion\newline}
 \end{minipage}
 \begin{minipage}[b]{0.4\linewidth}
  \centering
  \includegraphics[width=\linewidth]{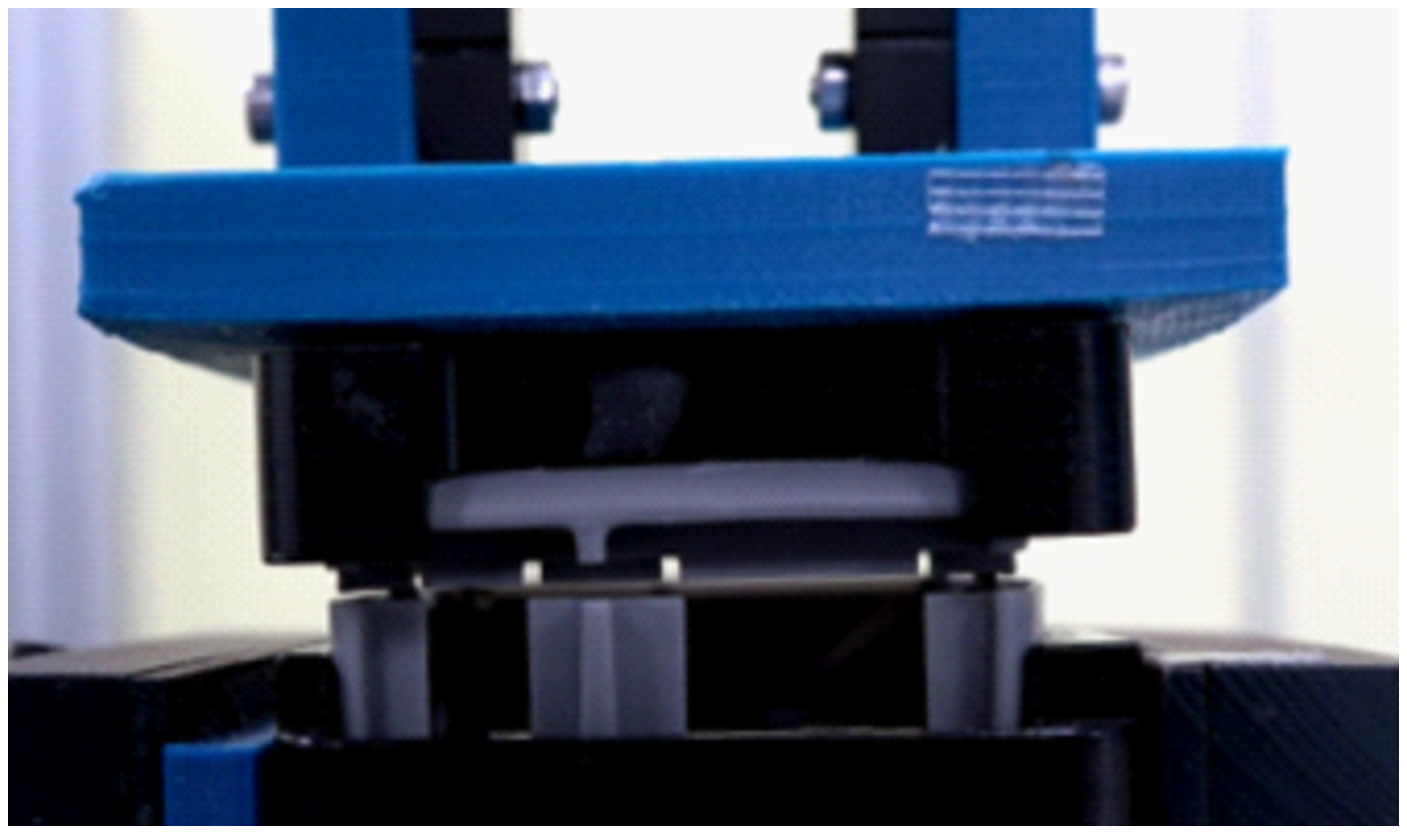}
  \subcaption{Additional probing in $+x$ direction}
 \end{minipage}\\
 \begin{minipage}[b]{0.4\linewidth}
  \centering
  \includegraphics[width=\linewidth]{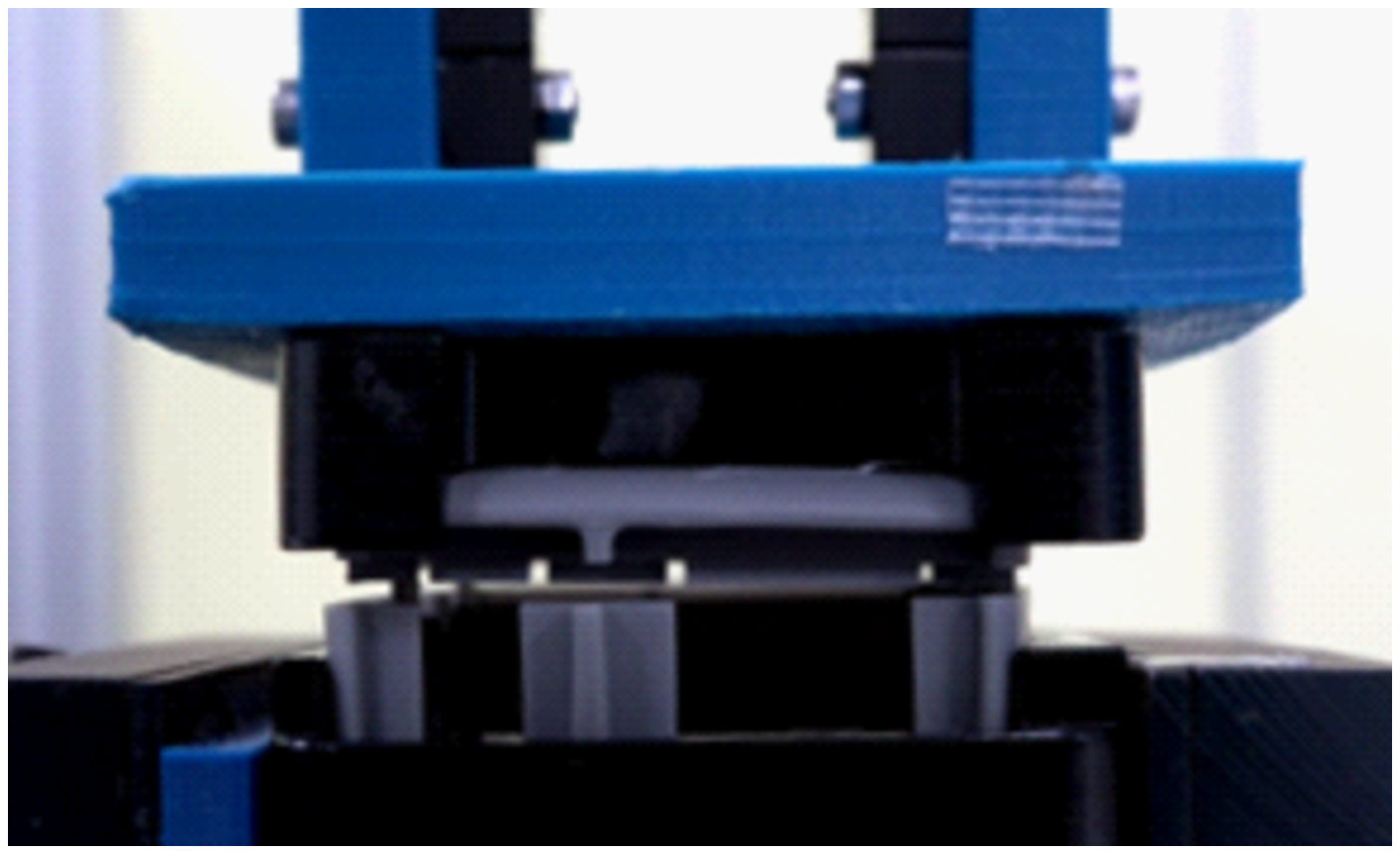}
  \subcaption{Additional probing in $-x$ direction}
 \end{minipage}
 \begin{minipage}[b]{0.4\linewidth}
  \centering
  \includegraphics[width=\linewidth]{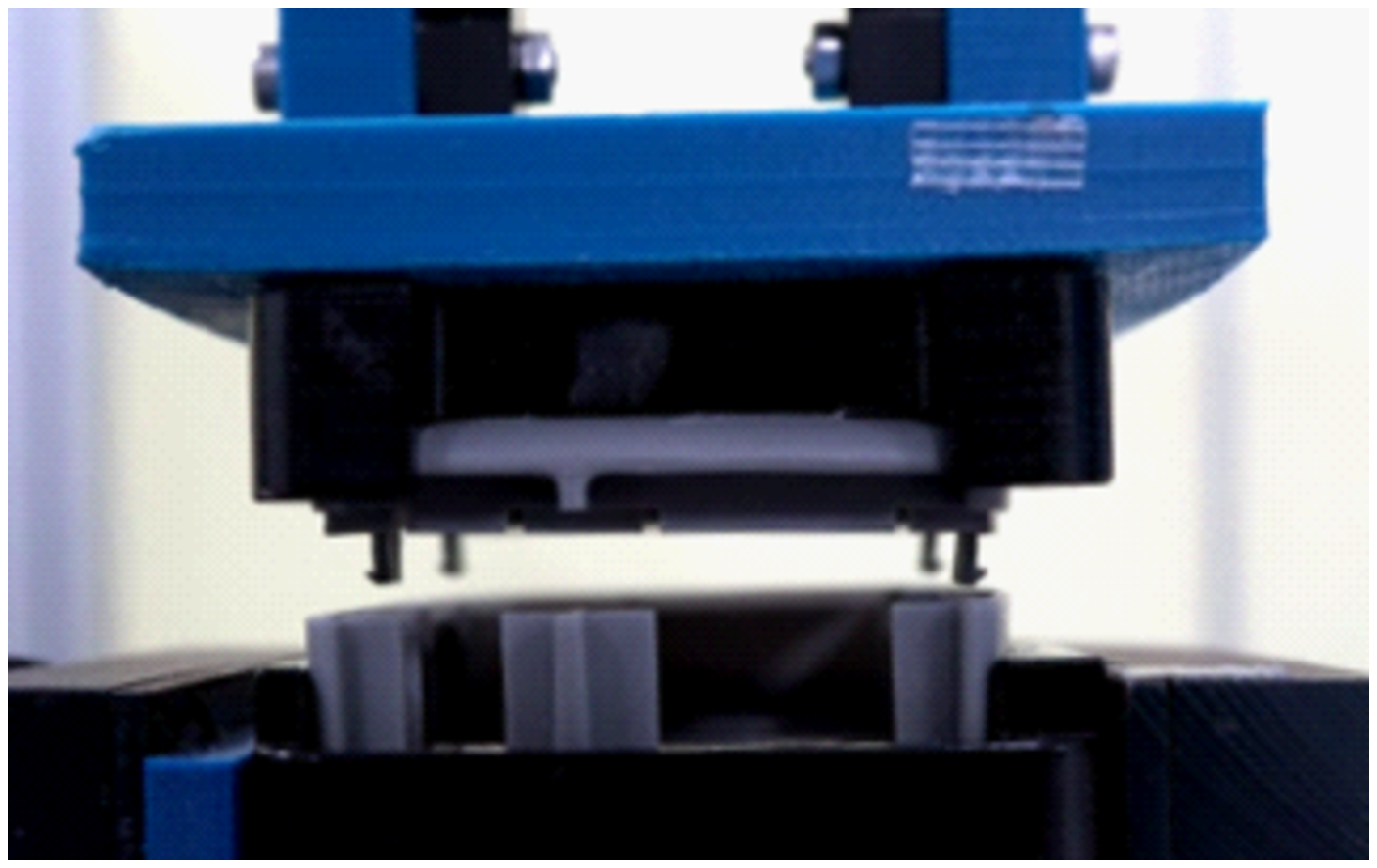}
  \subcaption{Error recovery motion\newline}
 \end{minipage}\\
 \begin{minipage}[b]{0.4\linewidth}
  \centering
  \includegraphics[width=\linewidth]{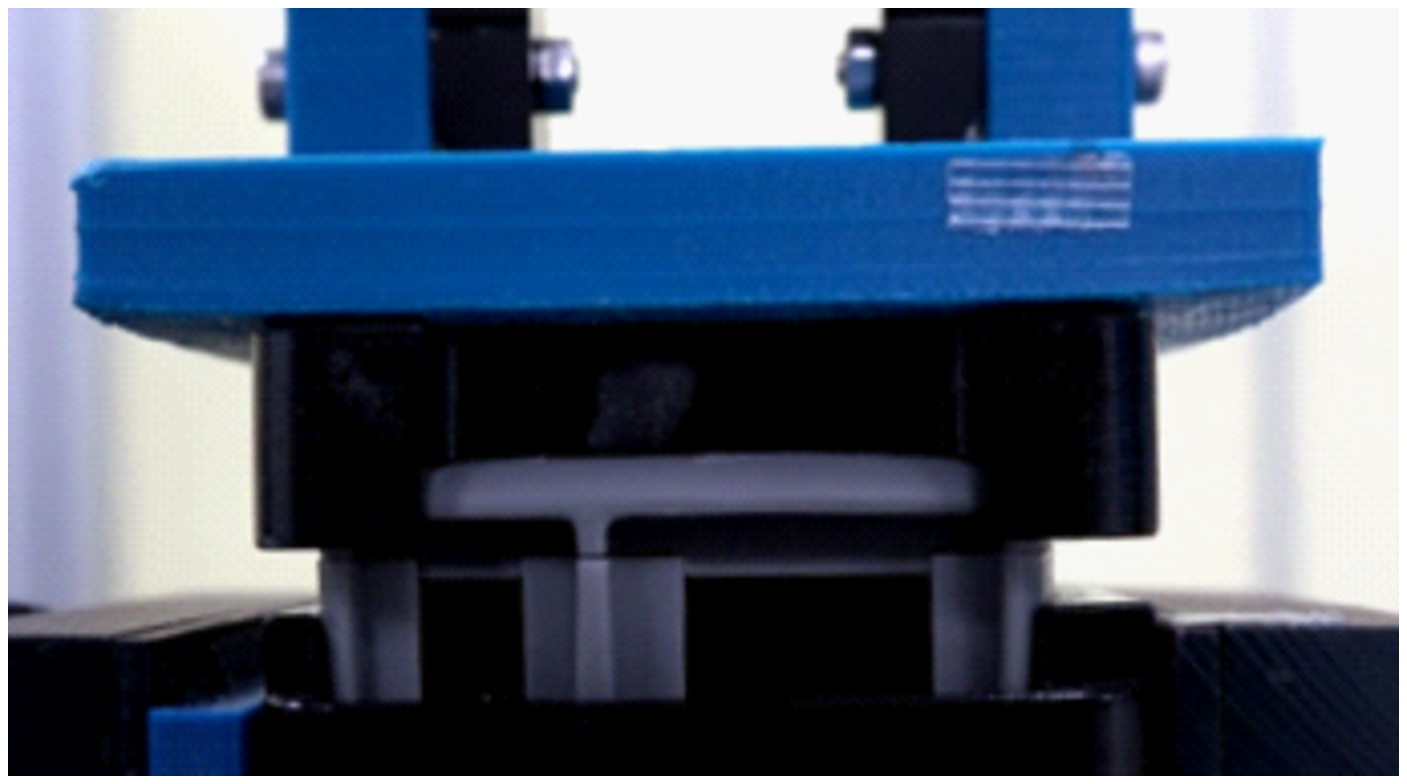}
  \subcaption{Snap assembly has successfully finished}
 \end{minipage}
 \caption{Error recovery from the offset pattern (3)}
    \label{recovery failure search}
    }
\end{figure}

We implemented the recovery motion from the identified error state. 
In this experiment, we tested the error recovery three times each for the following three cases: 

\begin{itemize}
    \item [(a)] After identified as the successful case (1), a robot further moves the male part in the $+z$ direction, and the task ($\Delta x= 0.5[{\mathrm{mm}}]$, $\Delta \theta_{z}=0.5$[deg]) is completed. Apparently, a robot does not need to recover from the error state in this case. 
    \item[(b)] After an error state is identified, a robot attempts to recover from the error state($\Delta x= 1.5[{\mathrm{mm}}]$, $\Delta \theta_{z}=-1.5$[deg]). 
    \item[(c)] After an error state is identified, a robot performs the additional probing. Then, a robot attempts to recover from the error state($\Delta x= -1.5[{\mathrm{mm}}]$, $\Delta \theta_{z}=-1.5$[deg]). 
\end{itemize}
Among three cases, a robot recovers from the identified error states in the cases (b) and (c). The snapshots of error recovery motion are shown in Figs. \ref{recovery success}, 
\ref{recovery failure nosearch} and \ref{recovery failure search} corresponding to the cases (a), (b) and (c), respectively. We tried the error recovery for three times for each case. The robot could successfully recover from the error state 
in all the cases without breaking the part, as shown in Table \ref{recovery table}. 

% \ref{4.5}節で正しく識別されたいくつかのオフセットパターンに対して，\ref{3.5}節に従ってエラーリカバリシステムを構築した．
% 本実験では，以下のエラーリカバリ動作の生成パターンに該当するオフセットパターン(1)$\sim$(3)について，それぞれ3回ずつエラーリカバリ動作を行った．\\
% (1)成否パターン(1)において，エラーリカバリ動作を行わずに，ロボットハンドを下($+z$)方向に降ろすことで作業を完了させる場合($\Delta x= 0.5[{\mathrm{mm}}]$，$\Delta \theta_{z}=0.5[^\circ]$)\\
% (2)作業が失敗パターンである場合に，探り動作工程に移行することなくエラーリカバリ動作を生成する場合($\Delta x= 1.5[{\mathrm{mm}}]$，$\Delta \theta_{z}=-1.5[^\circ]$)\\
% (3)作業が失敗パターンである場合に，探り動作工程に移行し，探り動作工程での識別結果からエラーリカバリ動作を生成する場合($\Delta x= -1.5[{\mathrm{mm}}]$，$\Delta \theta_{z}=-1.5[^\circ]$)\\
% エラーリカバリを行った様子を
% 図\ref{recovery success}，
% \ref{recovery failure nosearch}，
% \ref{recovery failure search}
% に，エラーリカバリの成功率を表\ref{recovery table}に示す．
% ここで，エラーリカバリの成功率については，成否パターンの識別結果が成功パターン，失敗パターンいずれにおいても，識別後にスナップアセンブリが正しく完了された場合に成功したとする．

\begin{table}[t]
\begin{center}
\caption{Result of error recovery}
  \begin{tabular}{|c|c|c|} \hline
  Offset pattern & Number of successful case & Success rate \\ \hline \hline
  (1) & 3 & 100 \\ \cline{1-3}
  (2) & 3 & 100 \\ \cline{1-3}
  (3) & 3 & 100 \\ \hline
  \end{tabular}
  \label{recovery table}
\end{center}
\end{table}

%From these results, a robot successfully recover from the error state without breaking the part. 
% 表\ref{recovery table}より，
% 全てのエラーリカバリ行動において，部品を破損することなく
% スナップアセンブリを完了することができた．
% このことから，エラーリカバリシステムの有用性が示唆された．

\color{black}
\section{conclusion}
\label{sec:conclusion}

In this paper, we proposed a method of predicting error states before the error actually occurs in robotic snap assembly for recovering from the identified error state. 
We perform the functional principal component analysis (fPCA) of the 6D force/torque profile. 
We confirmed that an error state is correctly identified by applying the obtained feature vector to the decision tree. We also confirmed that, if the estimation accuracy is low, we could better  identify the error state by additional probing. After identifying the error state, a robot successfully attempted to recover from the identified error state. 

In this paper, we confirmed the effectiveness of our approach by using a plastic parts with four snap joints. Application of our proposed method to other parts with different shape is considered to be our future research topic. In addition, we assumed just $\Delta x$ and $\Delta \theta_z$ to define the error states. As we increase the number of error states, it will become difficult to correctly estimate the error states. In a future research, we would increase the number of error states and see the estimation accuracy. The accuracy of estimation might depends also on the dynamics of the parts. Consideration on the dynamic effect is considered to be our future research topic.

% 本稿では，ロボットによるスナップアセンブリ工程において，従来では困難であるとされていた，失敗パターンを含めた作業結果の識別を作業途中で行うような成否パターン識別手法を提案した．さらに，識別されたエラーパターンに基づいて適切なエラーリカバリ行動を生成し，ロボットが自律的に作業に復帰するようなエラーリカバリシステムを構築した．
% 提案手法は，嵌め合い対象物体に様々なオフセットパターンを与えてスナップアセンブリを行い，力・トルクの波形データを取得する．
% そして，波形データ区間を変更しながら関数主成分分析を実行して特徴量分布を抽出し，
% SVMにより成否パターンの決定木を構築する．そして，決定木を参照することで，未知のスナップアセンブリの成否パターンを作業途中で識別する．
% 最後に，識別されたエラーパターンに基づいて，エラーを解消するようにロボットハンドを移動させることでエラーリカバリシステムを構築する．
% %分類精度と作業途中での識別精度がともに高いデータ区間，及び決定木を選択する．
% 実機実験の結果，
% 検証した未知のスナップアセンブリに対して作業途中での識別が達成され，提案手法の有用性が示唆された．
% エラーリカバリ行動の生成については，すべての試行において部品やロボットハンドが破損することなく作業への復帰が達成され，エラーリカバリシステムの有用性が示唆された．
% また本提案手法は，アセンブリ中にロボットハンドが取得する力情報のみを利用した手法であるため，スナップアセンブリ以外の組立作業工程においても有用であると考えられる．
% 今後の展望として，
% 決定木の構築手法については，
% より分類精度の高いノードが選択されるようにアルゴリズムを改良することや，
% 分類精度と作業途中での識別速度を両立するような評価式を提案することで，
% より作業の早い段階での識別を達成することを考えている．
% さらに，より高い識別成功率を達成するために，
% 嵌め合い時の情報から探り動作内容を適応的に変化させることで識別精度を向上させることが考えられる．
% エラーリカバリ動作については，成否パターンのみではなく，
% ずれの距離や角度といったより詳細な情報まで識別可能にすることが考えられる．
% さらに，より多様な成否パターンを追加した場合の提案手法の有用性を検証する予定である．
\color{black}

% \section*{Acknowledgement}
% This work was supported in part by the NACHI-FUJIKOSHI Corporation.

%%%%%%%%%%%%%%%%%%%%%%%%%%%%%%%%%%%%%%%%%%%%%%%%%%%%%%%%%%%%%%%%%%%%%%%%%%%%%%%%
% ========
%  Reference
% ========

\bibliographystyle{IEEEtran}
\bibliography{main}

\end{document}